\documentclass{aero}

\usepackage{bm}

\usepackage{float}

\definecolor{newcolor}{rgb}{.8,.349,.1}

\usepackage{longtable}
\usepackage{array}   

\ADsetup{
title={Adapting Critic Match Loss Landscape Visualization to Off-policy Reinforcement Learning},
author={Jingyi Liu, Jian Guo\cor{}, and Eberhard Gill},
address={Delft University of Technology, Delft, 2629 HS, the Netherlands},
email={J.Guo@tudelft.nl},
abstract={This work extends an established critic match loss landscape visualization method from online to off-policy reinforcement learning (RL), aiming to reveal the optimization geometry behind critic learning. Off-policy RL differs from stepwise online actor–critic learning in its replay-based data flow and target computation. Based on these two structural differences, the critic match loss landscape visualization method is adapted to the Soft Actor–Critic (SAC) algorithm by aligning the loss evaluation with its batch-based data flow and target computation, using a fixed replay batch and precomputed critic targets from the selected policy. Critic parameters recorded during training are projected onto a principal component plane, where the critic match loss is evaluated to form a 3-D landscape with an overlaid 2-D optimization path. Applied to a spacecraft attitude control problem, the resulting landscapes are analyzed both qualitatively and quantitatively using sharpness, basin area, and local anisotropy metrics, together with temporal landscape snapshots. Comparisons between convergent SAC, divergent SAC, and divergent Action-Dependent Heuristic Dynamic Programming (ADHDP) cases reveal distinct geometric patterns and optimization behaviors under different algorithmic structures. The results demonstrate that the adapted critic match loss visualization framework serves as a geometric diagnostic tool for analyzing critic optimization dynamics in replay-based off-policy RL-based control problems.},
keywords={reinforcement learning \sep loss landscape \sep adaptive dynamic programming \sep soft actor critic \sep attitude control}
}

\makeatletter
\def\ps@plain{%
  \let\@oddhead\@empty
  \let\@evenhead\@empty
  \let\@oddfoot\@empty
  \let\@evenfoot\@empty
}
\def\ps@headings{%
  \let\@oddhead\@empty
  \let\@evenhead\@empty
  \let\@oddfoot\@empty
  \let\@evenfoot\@empty
}
\makeatother

\begin{document}

\maketitle  


\Nomenclature

\begin{longtable}{|P{.25\linewidth}|P{.6\linewidth}|}
\hline
$\zeta$ & Chosen center point in the landscape graph \\
\hline
$\boldsymbol{\delta}, \boldsymbol{\eta}$ & Selected directions to visualize the landscape \\
\hline
$\alpha, \beta$ & Distances from the center point along directions $\delta$ and $\eta$ \\
\hline
$f(\alpha, \beta)$ & Loss function evaluated at perturbed parameters $(\alpha, \beta)$ \\
\hline
$\mathcal{L}\!$ & Loss function operator for the visualization\\
\hline
$J(t)$ & Cost function at time step  \\
\hline
$r(t)$ & Reinforcement signal (reward) at time step $t$ \\
\hline
$\gamma$ & Discount factor for future rewards (dimensionless) \\
\hline
$e_c(t), e_t^{\text{TD}}$ & TD error for the critic network \\
\hline
$\mathbf{w}_c$ & Weight of critic network \\
\hline
$\mathbf{w}_a$ & Weight of actor network \\
\hline
$s_t, a_t$ & System state and action at time step $t$ \\
\hline
$\pi(a|s)$ & Policy mapping state $s$ to action $a$ \\
\hline
$\ d_\pi$ & State-action visitation distribution induced by policy $\pi$ \\
\hline
$J(\pi)$ & Entropy-augmented objective \\
\hline
$\mathbb{E}[\cdot]$ & Expectation operator \\
\hline
$\mathcal{H}(\pi(\cdot|s_t))$ & Entropy of the policy distribution at state $s_t$ \\
\hline
$-\mathbb{E}_{a_t\sim\pi}[\log \pi(a_t|s_t)]$ & Definition of the entropy term \\
\hline
$\alpha_{\mathrm{temp}}$ & Temperature coefficient in SAC \\
\hline
$\mathcal{D}$ & Replay buffer or fixed probe set of states \\
\hline
$Q_{\mathbf{w}_c}(s_t,a_t)$ & Q-value estimated by the current critic network \\
\hline
$\log \pi(a_{t+1}|s_{t+1})$ & Log-probability of the sampled action \\
\hline

$L^{*}$ & Critic match loss at the final critic parameters\\

\hline
$\tilde L(\alpha,\beta)$ & Normalized relative critic match loss  \\
\hline
$\Delta L(\alpha,\beta)$ & Relative critic match loss  \\
\hline

$\epsilon$ & Radius of local neighborhood used for sharpness evaluation \\

\hline

$\theta_{\mathrm{dir}}$ & Angular direction variable on the circle of radius $\epsilon$ used in sharpness evaluation \\
\hline
$\mathrm{Sharp}_{\epsilon}$ & Sharpness index  \\

\hline
$\rho$ & Loss threshold defining the low-loss basin region \\

\hline
$A_{\rho}$ & Basin area where the normalized loss remains below threshold $\rho$ \\

\hline
$H$ & Hessian matrix of the normalized loss on the PCA plane \\

\hline
$\lambda_{\max}, \lambda_{\min}$ & Largest and smallest eigenvalues of the Hessian matrix $H$ \\

\hline
$\kappa$ & Condition number of the Hessian matrix on the PCA plane \\

\hline
$\log \kappa$ & Logarithmic measure of local anisotropy of the loss landscape \\

\hline
$x$ & State vector in space attitude dynamic model \\
\hline
$u$ & Control input vector ($\mathrm{N\cdot m}$) \\
\hline

$\bar{M}, \bm{M}$ & Control torque applied to spacecraft attitude system ($\mathrm{N\cdot m}$) \\
\hline
$\hat{J}_{\mathrm{sc}}$ & Spacecraft inertia matrix ($\mathrm{kg\, m^2}$) \\
\hline
$\bm{q} = [q_0, q_1, q_2, q_3]^T$ & Unit quaternion  \\
\hline
$\bm{\omega}$ & Angular velocity vector of spacecraft ($\mathrm{rad/s}$) \\
\hline
$\bm{\omega}_{\mathrm{target}}$ & Desired angular velocity vector ($\mathrm{rad/s}$) \\
\hline
$\dot{\bm{q}}$ & Time derivative of quaternion (dimensionless per second, $\mathrm{s^{-1}}$)\\
\hline
$e_{\mathrm{att}}$ & Attitude error term ($1 - q_0^2$) \\
\hline
$e_{\mathrm{rate}}$ & Angular rate error $\|\bm{\omega}-\bm{\omega}_{\mathrm{target}}\|^2$ \\
\hline
$e_{\mathrm{torque}}$ & Control effort error $\|\bm{u}\|^2$ \\
\hline
$k_{\mathrm{att}}$ & Weighting coefficient for attitude error \\
\hline
$k_{\mathrm{rate}}$ & Weighting coefficient for angular rate error \\
\hline
$k_{\mathrm{torque}}$ & Weighting coefficient for control effort \\
\hline
$\otimes$ & Quaternion multiplication operator \\
\hline

$\theta$ & Angular displacement from upright vertical position for cart-pole\\
\hline
$\dot{\theta}$ & Angular velocity of the pole \\
\hline
$x$ & Horizontal displacement of the cart \\
\hline
$\dot{x}$ & Horizontal velocity of the cart \\
\hline
$a$ & Normalized control action in $[-1,1]$ \\
\hline
$f_{\mathrm{cp}}$ & Force applied to the cart \\
\hline
$F_{\max}$ & Maximum allowable force magnitude \\
\hline
$m_c$ & Mass of the cart \\
\hline
$m$ & Mass of the pole \\
\hline
$l$ & Half-length of the pole \\
\hline
$g$ & Gravitational acceleration \\
\hline
$\ddot{\theta}$ & Angular acceleration of the pole \\
\hline
$\ddot{x}$ & Horizontal acceleration of the cart \\
\hline
$c(s,a)$ & Instantaneous quadratic cost function \\
\hline
$x_{\max}$ & Maximum allowable cart displacement (track limit) \\
\hline
$w_{\theta}$ & Weighting coefficient for angular displacement \\
\hline
$w_{\dot{\theta}}$ & Weighting coefficient for angular velocity \\
\hline
$w_{x}$ & Weighting coefficient for cart displacement \\
\hline
$w_{\dot{x}}$ & Weighting coefficient for cart velocity \\
\hline
$w_{u}$ & Weighting coefficient for control effort \\
\hline
\end{longtable}

\section{Introduction}

Reinforcement learning has been a hot spot for research in recent years. While the research was initiated by the computer science community, it quickly spread its applications and related theory development also to other research communities.

With its wide applications in optimization~\cite{hu2025robust}, control~\cite{recht2019tour}, and planning~\cite{zhang2025attention}, reinforcement learning has shown potential in solving decision-making problems. However, reinforcement learning algorithms often suffer from limited generalization capability, resulting in unstable training performance and inconsistent control outcomes across different environments. As a result, researchers are concerned about the interpretation of the performance of the reinforcement learning algorithms, rather than only treating it as a black box~\cite{9194697}. Interpretation methods, such as decision trees \citep{bastani2018verifiable}, natural language explanations \citep {hayes2017improving}, and visualization methods \citep{zhang2018visual}, have been developed and used to interpret the algorithm's behavior in decision making, planning, and other aspects. Among them, the visualization method is a widely used method for its intuitiveness, readability, and ease of interpretation by readers. The loss landscape is an index that is often chosen by researchers to visualize the internal process during the training of reinforcement learning methods \citep{kostrikov2021offline}, and also the characteristics of the neural networks that are used for approximation in the learning methods \citep{li2018visualizing}. 

The authors previously developed a method for visualizing the critic-matching loss landscape to analyze actor–critic learning behavior. That work focused on online reinforcement learning and demonstrated how the visualization reflects critic optimization patterns under continuously updated state–action data. Unlike online algorithms, which interact with the environment and update parameters in a stepwise manner, off-policy reinforcement learning adopts a replay-based training regime with decoupled data usage and parameter updates. Although the agent continues to interact with the environment, critic learning is performed on mini-batches sampled from a replay buffer, and target values are computed using delayed or target networks rather than being directly tied to the most recent interaction. Because of these structural differences, the method cannot be directly applied to off-policy RL without further adaptation. Bridging this gap ensures that the visualized loss geometry remains meaningful under the batch-based learning characteristic of off-policy RL.

The present work adapts the critic match loss landscape visualization method to suit the off-policy RL setting, using the Soft Actor–Critic (SAC) algorithm~\citep{haarnoja2018soft} as a representative case. SAC is chosen due to its strong and stable performance in control tasks~\citep{sola2022simultaneous,meijkamprobust}.To accommodate the replay-based training structure of SAC, the visualization method is adapted in two main aspects. First, to address the data flow difference, a fixed batch of replay data is used instead of continuously collected online samples, and the critic weight recording frequency is defined at the scale of neural network updates rather than episode boundaries. Second, to adapt the critic match loss computation, the target values are precomputed and fixed based on the final policy, incorporating the twin-critic structure and entropy regularization of SAC. With these adaptations, critic parameters recorded during training are projected onto principal directions identified through principal component analysis (PCA), and the critic match loss is evaluated to construct a three-dimensional loss landscape with an overlaid two-dimensional optimization path. Applied to a spacecraft attitude control problem, the resulting loss landscapes are analyzed both qualitatively and quantitatively using sharpness, basin area, and local anisotropy metrics. Temporal landscape snapshots are further constructed to examine how critic geometry evolves during training and to identify potential instability events. The results show that, under the frozen-target surrogate objective, the proposed visualization method provides structural insight into critic optimization behavior in the off-policy RL setting, extending its applicability beyond online actor–critic methods. Comparisons among convergent SAC, divergent SAC, and Action-Dependent Heuristic Dynamic Programming (ADHDP) cases reveal how different algorithmic structures shape critic optimization geometry and its relation to control behavior.

This paper is organized as follows. Section 2 introduces the loss function visualization technique, outlines the critic match loss landscape method and describes the adaptations required for its use in off-policy reinforcement learning. The quantitative indices for analyzing loss landscapes are introduced. Section 3 shows the setting of the spacecraft attitude system used for demonstrating the control algorithm's critic match loss landscape. Section 4 presents the critic match loss landscape visualization results, including comparative analyses between SAC and ADHDP, temporal landscape snapshots, and a discussion of the interpretation scope of the method. Section 5 validates the proposed critic match loss landscape method using SAC control on the Cart–Pole benchmark as a case study. Section 6 concludes the paper.

\section{Critic Match Loss Landscape Visualization Method and Its Adaptation to off-policy RL}

In this section, the visualization method for interpreting online RL algorithms is introduced. The Soft Actor Critic (SAC) algorithm is also provided, which serves as an object to be interpreted using the visualization method. 

\subsection{Visualizing the loss function}
In reinforcement learning, neural networks are used to approximate functions, often with a large number of parameters. The loss landscape, which shows how the loss varies with these parameters, can reveal properties such as flatness, sharpness, local minima, and saddle points. However, the high dimensionality of the parameter space makes direct visualization difficult. To address this, the Contour Plots and Random Directions method is proposed \citep{li2018visualizing}. The idea is to select two directions, $\boldsymbol{\delta}$ and $\boldsymbol{\eta}$, and plot the loss over the 2-D plane they span. The resulting surface is a projection slice of the full loss landscape. The loss value in this plane is computed as

\begin{equation}
f(\alpha,\beta) = \mathcal{L}\!\left(\boldsymbol{\zeta}^* + \alpha\,\boldsymbol{\delta} + \beta\,\boldsymbol{\eta}\right),
\label{eq:contour-plots}
\end{equation}
where $\boldsymbol{\zeta}^*$ is the chosen center point. $\alpha$ and $\beta$ represent displacements along the two directions. $\mathcal{L}\!$  is the loss function. This reduces the multidimensional problem to a 3-D plot, with $(\alpha, \beta)$ on the horizontal plane and $f(\alpha, \beta)$ on the vertical axis, making the structure of the loss landscape explicit and easy to interpret.

\subsection{Soft Actor-Critic (SAC) Algorithm}

In previous work, the loss visualization technique has been proposed to interpret the performance of online reinforcement learning algorithms with an actor-critic structure. To further extend the proposed visualization method beyond online learning, this study applies it to the Soft Actor–Critic (SAC) algorithm, which represents an off-policy reinforcement learning framework known for its stable control performance.

\begin{figure}[htbp]
    \centering
    \includegraphics[width=\columnwidth]{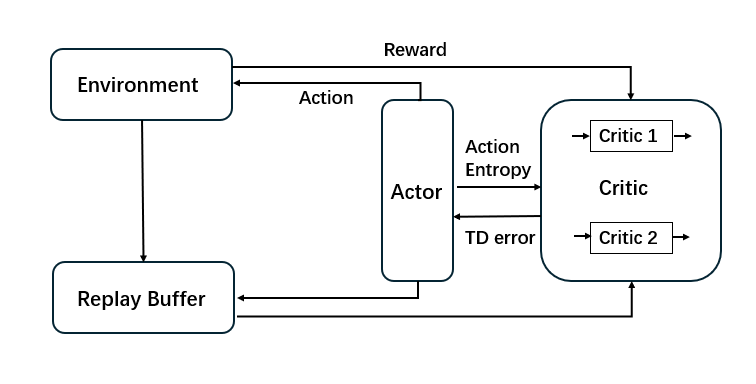}   
    \caption{SAC structure diagram}
    \label{fig:SAC_structure_diagram}
\end{figure}

~\autoref{fig:SAC_structure_diagram} shows the structure of the SAC algorithm. Similar to the ADHDP algorithm, it also employs an actor and critic structure. The actor is approximated to generate the actions, and the critic network is used to approximate the cost function. Different from the ADHDP algorithm, it uses two critic networks, one of which updates steps later than the other, to provide a better approximation of the cost function. 

Another factor where the SAC algorithm differs from the ADHDP algorithm is that the SAC algorithm is an off-policy reinforcement learning algorithm that optimizes a stochastic policy using an entropy-augmented objective function, which is the cost function in the case of the ADHDP algorithm. The goal of the Soft Actor–Critic (SAC) algorithm is to maximize both the expected cumulative reward and the entropy of the policy:
\begin{equation}
J(\pi) = \sum_{t=0}^{\infty} \mathbb{E}_{(s_t, a_t) \sim d_\pi} \left[ r(s_t, a_t) + \alpha \mathcal{H}(\pi(\cdot | s_t)) \right],
\end{equation}
where $\mathcal{H}(\pi(\cdot | s_t)) = -\mathbb{E}_{a_t \sim \pi}[\log \pi(a_t | s_t)]$ is the entropy term, and $\alpha > 0$ is the temperature coefficient that balances the trade-off between reward maximization and exploration.

The critic networks are trained to minimize the Bellman residual using the target Q-value:
\begin{equation}
J_Q(\mathbf{w}_c)
= \mathbb{E}_{(s_t, a_t, r_t, s_{t+1}) \sim \mathcal{D}} \left[ \left( Q_{\mathbf{w}_c}(s_t, a_t) - y_t \right)^2 \right],
\end{equation}
where the target value $y_t$ is computed as:
\begin{equation}
y_t = r_t + \gamma \, \mathbb{E}_{a_{t+1} \sim \pi} \left[ Q_{\bar{\mathbf{w}_c}}(s_{t+1}, a_{t+1}) - \alpha \log \pi(a_{t+1} | s_{t+1}) \right].
\label{eq:SAC_target}
\end{equation}

Here, $s_t$ and $a_t$ denote the system state and action at time step $t$, respectively, while $r_t$ is the immediate reward. The term $\ d_\pi$ represents the state–action distribution under policy $\pi$. The parameter $\mathbf{w}_c$ refers to the critic network weights, and $\bar{\mathbf{w}_c}$ denotes the slowly updated target critic network weights. The discount factor $\gamma \in (0,1)$ determines the importance of future rewards. The replay buffer $\mathcal{D}$ stores sampled transitions $(s_t, a_t, r_t, s_{t+1})$ that are used to train the critic network.

\subsection{Visualizing the critic match loss function for Soft Actor Critic algorithm}
\label{subsection: SAC loss landscape method}
The idea of visualizing the critic match loss function was previously developed by the authors to analyze the critic network behavior of online reinforcement learning algorithm. In that context, the visualization was shown to reflect critic optimization patterns under continuously updated state–action data. Besides online RL algorithms, off-policy reinforcement learning algorithms are also widely used in control scenarios. However, off-policy RL algorithms often have more complicated structures. So it's important to see how the established critic match loss landscape visualization method can be adapted to off-policy RL settings. In this work, the Soft Actor Critic Algorithm is selected as the sample algorithm for off-policy RL. 

Before introducing the adaptation of the critic match loss visualization method to off-policy reinforcement learning, the visualization procedure for online RL is briefly summarized. For an online actor–critic reinforcement learning algorithm, the critic network is employed to approximate the cost function, and its approximation accuracy directly affects the control performance. The temporal-difference (TD) error reflects this approximation accuracy and is iteratively minimized during critic training. However, both the TD error and the distribution of system states evolve continuously as training progresses, which makes it difficult to represent the training process using a single, consistent TD-error surface. To address this issue, the critic match loss is introduced as a surrogate measure to represent the evolution of the training signal in a fixed reference setting.

To capture the training evolution, the critic weight vector is recorded at the end of each training episode, forming a sequence of weight snapshots that reflects the progression of critic updates over the entire training process. Since the critic weights are high-dimensional, dimensionality reduction is required for visualization. As indicated in \autoref{eq:contour-plots}, two directions are selected to serve as the axes of the loss landscape. These directions are obtained by applying principal component analysis (PCA) to the set of recorded critic weight vectors $\mathbf{w}_c$, yielding two mutually orthogonal basis vectors. Denoted as $\boldsymbol{\delta}$ and $\boldsymbol{\eta}$, these directions define a low-dimensional parameter plane for visualizing variations in $\mathbf{w}_c$. Rather than depicting the loss along a single dimension, the loss surface is explored in two dimensions by sampling the parameters $\alpha$ and $\beta$ over a predefined grid. For each $(\alpha, \beta)$ pair, the critic weights are reconstructed as $\mathbf{w}_c' = \mathbf{w}_c + \alpha \delta + \beta \eta$, and the corresponding critic match loss is evaluated.

For each grid point, the loss is computed using fixed input–target pairs. The inputs consist of system states collected from the final training episode, while the targets are the corresponding temporal-difference targets computed at the final epoch. This procedure yields a three-dimensional loss landscape that characterizes the local loss geometry around the final policy. By projecting the loss landscape onto the same PCA plane, the recorded critic weight trajectory can be overlaid as a two-dimensional optimization path. This path provides a qualitative representation of how the critic parameters evolve relative to the underlying loss geometry during training. Although this construction does not reproduce the exact time-varying TD objective used throughout training, it offers an interpretable view of the critic’s optimization behavior and supports the analysis of convergence and instability in online reinforcement learning.

To adapt the critic match loss landscape technique from online to off-policy setting, it's necessary to compare the structure differences between online RL and off-policy RL, which will show the aspects the match loss landscape technique has to be adapted. 

1. Firstly, the data flow is different. Different from stepwise online actor-critic reinforcement learning, in which the sample data is collected per time step and the weight updates accordingly, SAC is an off-policy reinforcement learning algorithm with replay-based updates. Unlike strictly offline batch RL, the SAC implementation used in this work continues to interact with the environment throughout training. At each time step, the agent interacts with the environment and stores the resulting transition $(s_t, a_t, r_t, s_{t+1})$ into a replay buffer. Environment interaction continues throughout training and the replay buffer grows incrementally as new transitions are collected. Once the buffer contains enough samples to form a mini-batch, the neural networks are updated by repeatedly sampling batches from the buffer. And the weights of the neural networks update accordingly.

2. Secondly, the target calculation in TD error is different. TD error is used to calculate the critic match loss and generate the local geometry of the critic match loss function, which can show how the critic match loss varies in the neighborhood of the final policy. From \autoref{eq:SAC_target}, the TD error can be expressed as
\begin{equation}
\begin{aligned}
e_t^{\text{TD}}
&= Q_{\mathbf{w}_c}(s_t, a_t) \\
&\quad - \Bigl(
      r_t
      + \gamma \, \min_{i=1,2} Q_{\bar{\mathbf{w}_c}_i}(s_{t+1}, a_{t+1}) \\
&\qquad\qquad
      - \alpha \log \pi(a_{t+1} \mid s_{t+1})
    \Bigr)
\end{aligned}
\label{eq:SAC_TD_error}
\end{equation}

where $Q_{\mathbf{w}_c}(s_t, a_t)$ is the Q value of the current critic network, $Q_{\bar{\mathbf{w}_c}_i}$ is the current target network which uses soft update,$r_t$ is the current reward, $\gamma$ is the discount factor,  $\pi(a_{t+1} \mid s_{t+1})$ is the probability density of action $a_{t+1}$ under the state $s_{t+1}$, $\alpha$ is the temperature coefficient that controls the trade-off between policy exploration and exploitation and  $a_{t+1} \sim \pi(\cdot \mid s_{t+1})$ is the next action sampled under the current policy with state $s_{t+1}$. For online RL, TD error expression is much more simple, which is
\begin{eqnarray}
e_c(t)=[r(t)+\gamma J(t)]-J(t-1)
\label{eq:online TD Error}
\end{eqnarray}
The difference in the target computation in TD error will determine the calculation of critic match loss.

From the discussion above, it can be seen that the structural differences between online and off-policy reinforcement learning mainly lie in (i) the data usage flow and (ii) the target computation mechanism for critic updates. Accordingly, the visualization method is adapted along two directions: adaptation to the data flow and adaptation to the critic match loss computation.

\textbf{A. Adaptation to the off-policy data flow}

To adapt to the data flow in off-policy RL, two design choices are introduced.

(1) Using a fixed batch of replay data instead of continuously collected online samples.
In online RL, the sample distribution evolves as the policy changes and new trajectories are generated. For online RL, the states and loss from the last training episode are used for the critic match loss of each grid. Using the data from the last training episode, the final policy is evaluated through the critic loss landscape.  In off-policy RL, critic training is batch-based and driven by replay samples. Therefore, a fixed batch is sampled from the replay buffer, which is consistent with SAC's training mechanism. 

(2) Recording the critic weights at a frequency defined by network update steps.
For an online reinforcement learning method, the weight recording frequency is selected based on the real-world dynamics. For the online RL visualization, the weight is recorded when each episode ends, which either meets the simulation time limit or system states limit. However, as described above, the data usage flow for neural networks and the weight update flow is different in the case of an off-policy reinforcement learning algorithm. SAC updates the networks by repeatedly sampling mini-batches from the replay buffer, and a large number of parameter updates may occur within a single episode. As a result, the weight recording frequency is chosen on the same scale as the neural network update steps. Specifically, the network parameters are recorded at fixed intervals during training. These recorded parameter sets are then used to project the loss landscape over the entire training process.

\textbf{B. Adaptation to the critic match loss computation in SAC}

A fixed batch of state ``$s_t$, $a_t$, $r_t$, $s_{t+1}$'' samples are collected to calculate the corresponding Bellman targets $y_t$ using \autoref{eq:SAC_target}, where the expectation term over the next action is evaluated using the final policy. To adapt to this critic match loss computation scheme, three design choices are introduced.

(1) SAC Bellman targets are precomputed and fixed for a selected batch of replay data. 
In SAC,  the target value depends on the twin-critic structure, the entropy term, and policy sampling. During visualization, the critic parameters are systematically perturbed on a weight grid that does not correspond to actual training states. Therefore, recomputing targets for each grid point would introduce ambiguity and make the loss definition inconsistent. Fixing the targets computed from the final policy provides a well-defined reference for evaluating the critic loss across the weight grid. 

(2) The visualization focuses on the primary critic network within the twin-critic architecture. 
Both critics are trained using the same loss formulation and share an identical target definition. Therefore, focusing the visualization on a single critic network does not alter the SAC training objective and allows the loss landscape to be defined consistently.

(3) The actor network and the temperature parameter are kept fixed during visualization. 
Since the goal is to examine how the critic loss varies with respect to critic parameters alone, allowing the actor or entropy coefficient to change would couple multiple sources of variation. Freezing these components isolates the local loss geometry of the critic while remaining consistent with the final stage of SAC training.

From the method description above, there are two main adaptions when the critic match loss visualization is applied to off-policy actor-critic RL algorithm. The first main adaption is caused by the data flow difference between online and off-policy RL. For the adaption in data flow, two design choices are made (1) using a fixed batch of replay data instead of continuously collected online samples; (2) critic weight recording frequency. For the adaption in calculation critic match loss, three design choices are made. (1) precomputing and fixing the target values that combine the twin-critic structure and the entropy term; (2) focusing the analysis on the primary critic within the twin-critic pair; and (3) freezing the actor and temperature parameters to isolate the critic’s local loss geometry. These modifications align the visualization method with the SAC training mechanism while preserving its original interpretive purpose, enabling consistent comparison between online and off-policy RL paradigms.

The above description of constructing critic match loss landscape is illustrated using the final trained policy. Yet the formulation is not restricted to the terminal stage of training. By recording the critic parameters, associated target quantities, and a representative replay batch at any selected time point, and freezing these components, a locally stationary surrogate objective can be defined for that stage. Training then proceeds with further environment interaction and buffer growth. The resulting landscape therefore characterizes the critic geometry corresponding to a specific policy snapshot. By constructing such frozen-target landscapes at multiple training stages, the method enables a temporal analysis of critic optimization behavior. While each surface remains a static approximation, the sequence of snapshots provides insight into how critic stability and geometric structure evolve throughout learning.

By projecting the 3-D landscape onto the 2-D PCA plane obtained from the recorded critic weights, the optimization trajectory of the critic parameters during SAC training can be visualized on the same contour map. This 2-D path qualitatively represents how the weights evolve in relation to the underlying loss geometry, thereby revealing the optimization trend and convergence behavior throughout the training process.

This procedure constructs a stationary surrogate of the critic’s training objective under a selected policy and a qualitative representation of the projected optimization path. Although it does not reproduce the exact stepwise temporal difference objectives used during SAC training, it enables the visualization of how the critic match loss varies in the final policy, providing a clear depiction of the local geometry that characterizes the optimization behavior of the critic of SAC.

\subsection{Quantitative analysis of loss landscape}
\label{subsection: loss landscape quantity analysis}
The critic match loss landscape is derived using the method in \autoref{subsection: SAC loss landscape method}. It will directly show the qualitative representation of the optimization path. To further enable objective comparison and interpretation beyond visual inspection, a quantitative analysis of the loss landscape is introduced here.

Three indices are considered in this work: \emph{sharpness}, \emph{basin area}, and \emph{local anisotropy}. These indices capture different aspects of the loss geometry around the final critic parameters and provide complementary
descriptions of the training outcome.

Since the magnitude of the critic match loss varies significantly across algorithms and control systems, a direct comparison of raw loss values is not meaningful. Therefore, the loss landscape is first expressed relative to the final critic parameters. Let $L(\alpha,\beta)$ denote the critic match loss on the PCA plane, and let $(\alpha^{*},\beta^{*})$ be the final critic parameters with loss value $L^{*}$. A relative loss surface is defined as
\begin{equation}
\Delta L(\alpha,\beta) = L(\alpha,\beta) - L^{*}.
\end{equation}
To remove scale dependence, $\Delta L$ is normalized by a robust internal scale, chosen as the inter quartile range (IQR) of $\Delta L$ over the grid. The resulting dimensionless surface $\tilde L$ enables consistent geometric interpretation across different training runs.

\paragraph{Sharpness}
Sharpness describes how rapidly the loss increases when the critic parameters are perturbed away from the final point. For a given radius $\epsilon$, it is computed as 
\begin{equation}
\mathrm{Sharp}_{\epsilon} =
\max_{\theta_{\mathrm{dir}}\in[0,2\pi)}
\tilde L\!\left(
\alpha+\epsilon\cos\theta_{\mathrm{dir}},\;
\beta+\epsilon\sin\theta_{\mathrm{dir}}
\right).
\end{equation}
This index represents the worst-case local sensitivity of the critic loss within a neighborhood of radius $\epsilon$. $\theta_{\mathrm{dir}}$ denotes the angular direction on the circle of radius $\epsilon$ and it is unrelated to the network parameter vector.

From the perspective of reinforcement learning evolution, a larger sharpness indicates that the final critic parameters lie in a region where deviations are strongly penalized, suggesting a locally well-defined solution with a clear descent structure. In contrast, a small sharpness value corresponds to a flat or weakly constrained region, which is often associated with unstable or non-convergent training behavior.

\paragraph{Basin area}
The basin area characterizes the spatial extent of low-loss regions surrounding the final critic parameters. It is defined as the area of the region where the normalized loss remains below a prescribed threshold $\rho$,
\begin{equation}
A_\rho =
\mathrm{Area}\left\{
(\alpha,\beta)\;|\;
\tilde L(\alpha,\beta)\le\rho
\right\}.
\end{equation} 
This quantity provides a measure of how broadly the critic can vary while maintaining a comparable loss level.

In the context of reinforcement learning, a larger basin area suggests that the training process has converged to a solution that is robust to parameter perturbations, whereas a very small basin indicates a fragile solution that
depends on fine parameter tuning. It should be noted that when training fails to converge, the loss landscape may not exhibit a closed basin structure, in which case this value does not convey meaningful information.

\paragraph{Local anisotropy}
Local anisotropy describes the directional imbalance of the loss landscape near the final critic parameters. A quadratic approximation of $\tilde L$ is constructed in a small neighborhood of $(\alpha^{*},\beta^{*})$, yielding a
$2\times2$ Hessian matrix $H$ on the PCA plane. The degree of anisotropy is measured by the logarithm of the condition number, 
\begin{equation}
\log\kappa = \log\!\left(\frac{\lambda_{\max}(H)}{\lambda_{\min}(H)}\right),
\end{equation}
where $\lambda_{\max}$ and $\lambda_{\min}$ are the largest and smallest
eigenvalues of $H$, respectively.The condition number of the resulting Hessian matrix on the PCA plane captures the relative difference between steep and flat directions of the loss. 

From an optimization viewpoint, strong anisotropy indicates an  ill-conditioned valley, where progress is sensitive to step size and update direction. In reinforcement learning, such landscapes are often associated with
slow convergence or instability, whereas more isotropic curvature corresponds to smoother and more reliable training dynamics.

Together, these indices provide a compact quantitative description of the critic loss landscape. When combined with the visualization of the optimization path, they offer insight into how different reinforcement learning algorithms evolve in parameter space and why their training outcomes may differ.

\section{Spacecraft Attitude Control Setting}
\label{section: SAC attitude control setting}
In this section, the spacecraft attitude system is selected as the sample system for SAC control. The control results of using the SAC  algorithm are shown.

In this work, the combined spacecraft in the ADR scenario will be used as the control object for the SAC algorithm. From \citep{Shan2016}, the combined spacecraft in an ADR mission is a system affected by different kinds of uncertainty. In this work, to test the control algorithm for the post-capture stage in a gradual way, the simplest case is studied first, where the only uncertainty comes from the inertial parameters. The following assumptions are made. 
\begin{itemize}
  \item The target is firmly grasped by rigid robotic manipulators mounted on a rigid servicing spacecraft.
  \item After capture, all manipulator joints are locked \citep{huang2018postcapture}.
  \item The target is rigid, uncooperative, and has no control capability.
\end{itemize}

With these conditions, the combined spacecraft is treated as a single rigid body with unknown inertial parameters. The dynamic model of the spacecraft attitude system is described in detail in Appendix ~\ref{app:Appendix attitude dynamics}.

Unless otherwise specified, the general simulation settings for the spacecraft attitude control experiments are summarized below. The spacecraft attitude control policy was trained using the Soft Actor--Critic (SAC) algorithm from the Stable-Baselines3 library \citep{stable-baselines3}. The simulation environment was the quaternion-based attitude dynamics model described in Appendix ~\ref{app:Appendix attitude dynamics}. The SAC agent used the default multi-layer perceptron (MLP) policy network. The training was run for a total of $2\times 10^5$ time steps, with the simulation time step set to $0.02\,\text{s}$.  Two Q-networks were used, each with two hidden layers of $256$ units and ReLU activation. The policy was evaluated every few thousand steps during training to ensure stable learning. After training, the final policy was tested in a deterministic rollout of $500$ simulation steps, and the resulting quaternion, angular velocity, and control torque histories were recorded for analysis.

With the attitude dynamics model introduced in Appendix ~\ref{app:Appendix attitude dynamics}, 
the spacecraft was modeled as a rigid body with inertia matrix $\mathbf{J} = \mathrm{diag}(3.0,\, 2.0,\, 1.0)~\mathrm{kg \cdot m^2}$, 
and the simulation used a fixed integration step size $\Delta t = 0.02~\mathrm{s}$. 
The target attitude was the unit quaternion $\mathbf{q}_{\mathrm{target}} = [1,\, 0,\, 0,\, 0]^\mathsf{T}$, corresponding to zero rotation, 
and the target angular velocity was $\bm{\omega}_{\mathrm{target}} = \mathbf{0}~\mathrm{rad/s}$. 
The initial state was set to a small rotation of $0.2~\mathrm{rad}$ about each Euler axis, with an initial angular velocity $\bm{\omega}(0) = [0.1,\, 0.2,\, -0.1]^\mathsf{T}~\mathrm{rad/s}$. 

The control torque was bounded by $\|\bm{u}\|_\infty \le 5.0~\mathrm{N \cdot m}$. 
The reward function and its error terms were defined as
\begin{subequations}\label{eq:reward_errors}
\begin{align}
r             & = -\bigl( k_{\mathrm{att}}\, e_{\mathrm{att}}
                     + k_{\mathrm{rate}}\, e_{\mathrm{rate}}
                     + k_{\mathrm{torque}}\, e_{\mathrm{torque}} \bigr), \\
e_{\mathrm{att}}    & = 1 - q_0^2, \\
e_{\mathrm{rate}}   & = \left\| \bm{\omega} - \bm{\omega}_{\mathrm{target}} \right\|^2, \\
e_{\mathrm{torque}} & = \left\| \bm{u} \right\|^2.
\end{align}
\end{subequations}
where $r$ denotes the instantaneous reward, and $e_{\mathrm{att}}$, $e_{\mathrm{rate}}$, and $e_{\mathrm{torque}}$ represent the attitude, angular rate, and control effort errors, respectively. 
The weighting coefficients $k_{\mathrm{att}} = 10.0$, $k_{\mathrm{rate}} = 1.0$, and $k_{\mathrm{torque}} = 0.1$ determine the relative importance of each error term in the total reward. 
Here, $q_0$ is the scalar part of the unit quaternion $\bm{q} = [q_0, q_1, q_2, q_3]^{\mathrm{T}}$ representing the spacecraft attitude, $\bm{\omega}$ is the body angular velocity vector, $\bm{\omega}_{\mathrm{target}}$ is the desired angular velocity, which is set to zero for attitude stabilization control, and $\bm{u}$ is the control torque vector applied to the spacecraft. This setup was chosen to encourage accurate attitude tracking with minimal rate deviation and control usage.

    


\section{Critic Match Loss Landscape Visualization Results}

This section presents the critic match loss landscape analysis across multiple control scenarios to examine its interpretive capability and methodological scope. We first demonstrate the landscape corresponding to a convergent SAC control case, followed by a comparison between convergent SAC and divergent ADHDP to highlight structural differences under distinct learning paradigms. Next, a divergent SAC case is analyzed together with temporal landscape snapshots to investigate instability events during training. A direct comparison between convergent and divergent SAC cases is then conducted to examine how critic optimization trajectories differ under successful and unsuccessful control outcomes. Finally, we clarify the interpretation scope and methodological boundaries of the proposed visualization framework as a geometric diagnostic tool for critic optimization behavior.

\subsection{Final Critic Match Loss Landscape of convergent SAC control}
\label{subsection:convergent_SAC}

For the simulation of convergent SAC control, the main hyperparameters followed the default values in Stable-Baselines3: a learning rate of $3\times 10^{-4}$, a replay buffer size of $10^6$, a batch size of $256$, $\gamma = 0.99$, $\tau = 0.005$, and automatic entropy tuning enabled. Actor and critic networks were updated at every step with 1 gradient iteration.

\begin{figure*}[htbp]
    \centering
    \subfigure[SAC control result]{
        \centering
        \includegraphics[width=0.6\textwidth]{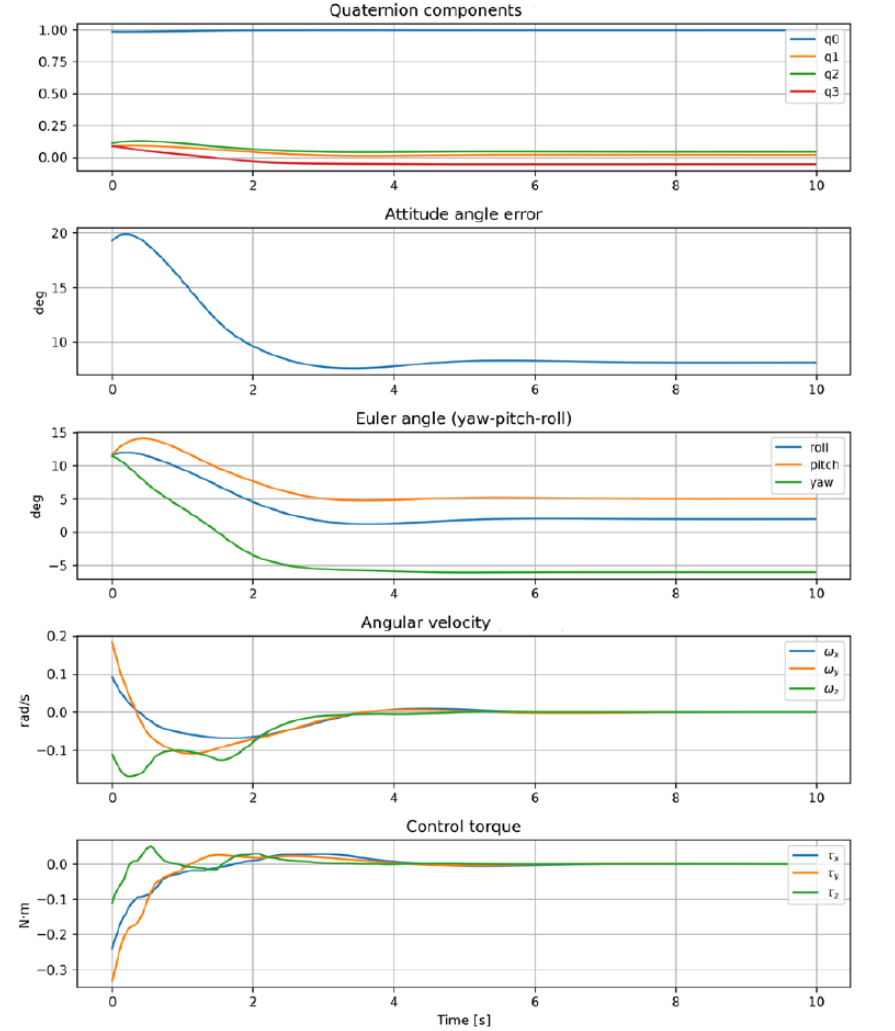}
        \label{fig: SAC system performance}
}
    \hspace{0.02\columnwidth}
    \subfigure[SAC training curve]{
        \centering
        \includegraphics[width=0.6\textwidth]{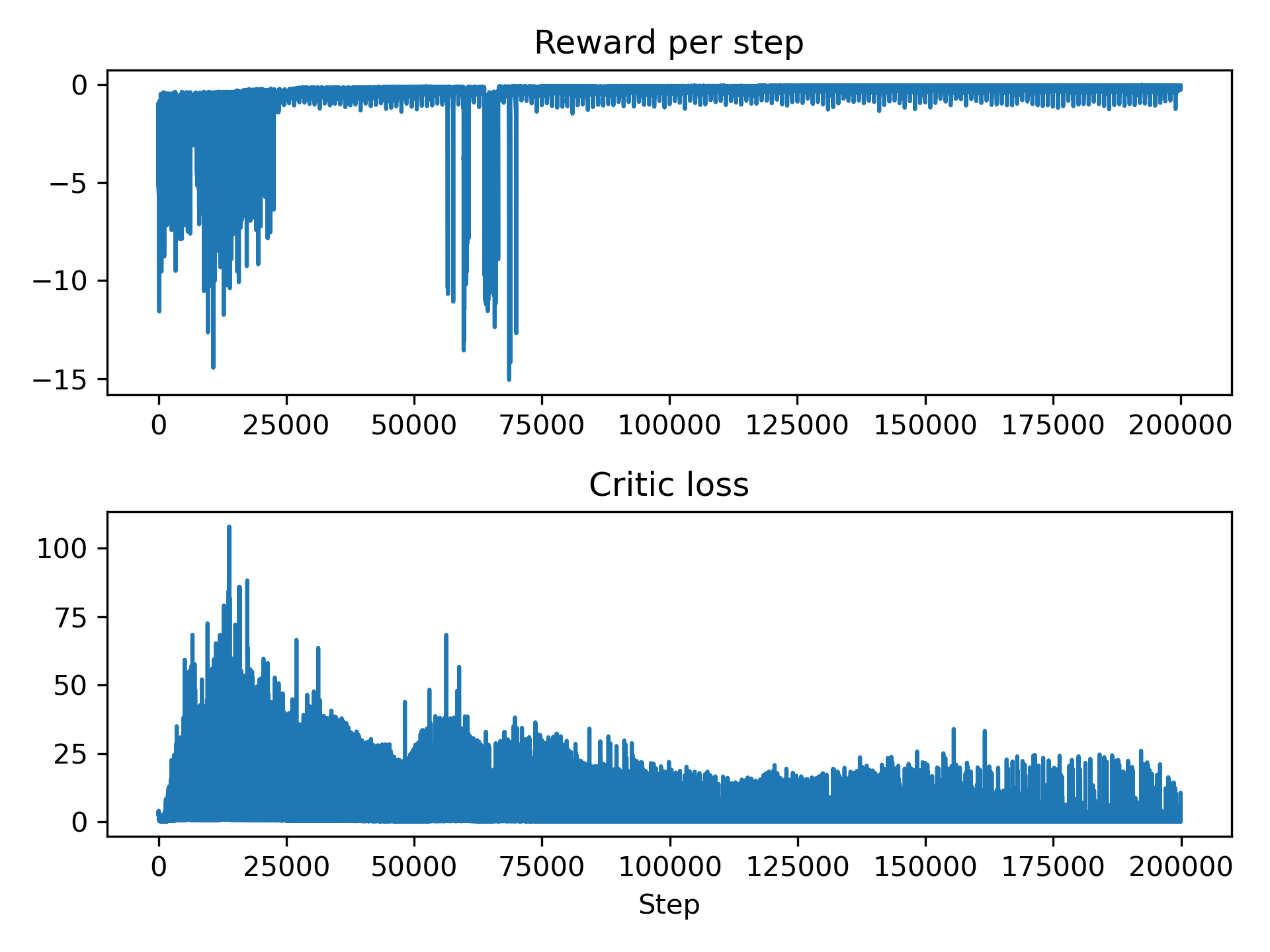}
        \label{fig: SAC training curve}
}

    \caption{SAC convergent control results for spacecraft attitude system}
    \label{fig:SAC convergent sc system and algorithm performance}
\end{figure*}

\autoref{fig:SAC convergent sc system and algorithm performance} shows the spacecraft system performance and the training process of the SAC algorithm. From \autoref{fig: SAC system performance}, it can be seen that the quaternions are stabilized around the given control target, which is $[1,\, 0,\, 0,\, 0]^\mathsf{T}$. Under the given control torque limit, the angular velocity is decreased from the initial state to around zero. These results confirm that the SAC policy successfully achieves stable spacecraft attitude control.

From \autoref{fig: SAC training curve}, it can be seen that the reward is accumulated through training. The critic loss, which is initially large and fluctuating, gradually decreases and stabilizes, reflecting the convergence of Q-value estimation under the entropy-regularized objective of SAC. The training behavior demonstrates effective policy learning and critic approximation during training.

\begin{figure}[htbp]
    \centering
    \subfigure[SAC 3-D loss landscape]{
        \centering
        \includegraphics[width=0.5\textwidth]{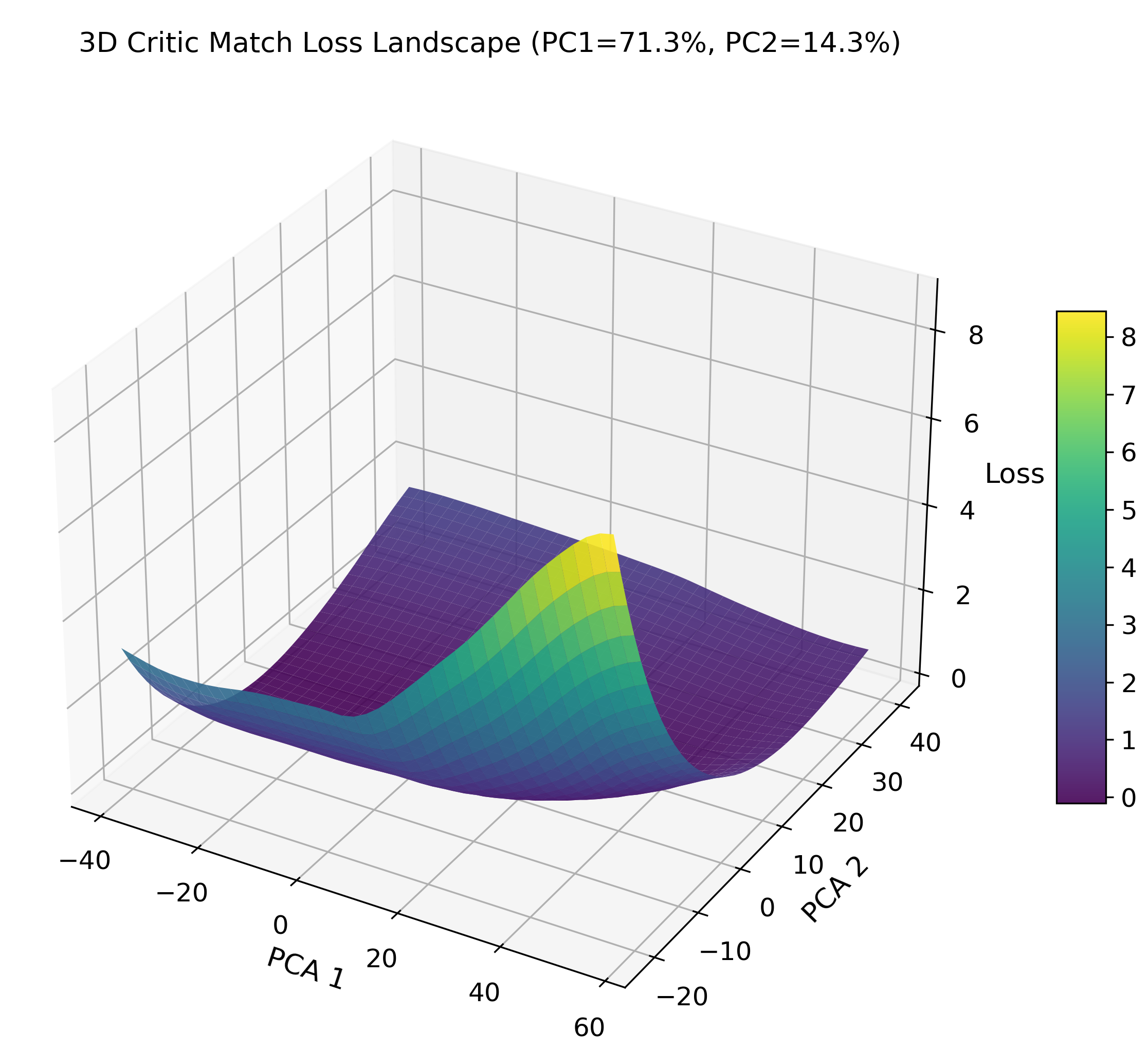}
        \label{fig:SAC_3D loss}
    }
    \hspace{0.02\columnwidth}
    \subfigure[SAC 2-D loss curve]{
        \centering
        \includegraphics[width=0.5\textwidth]{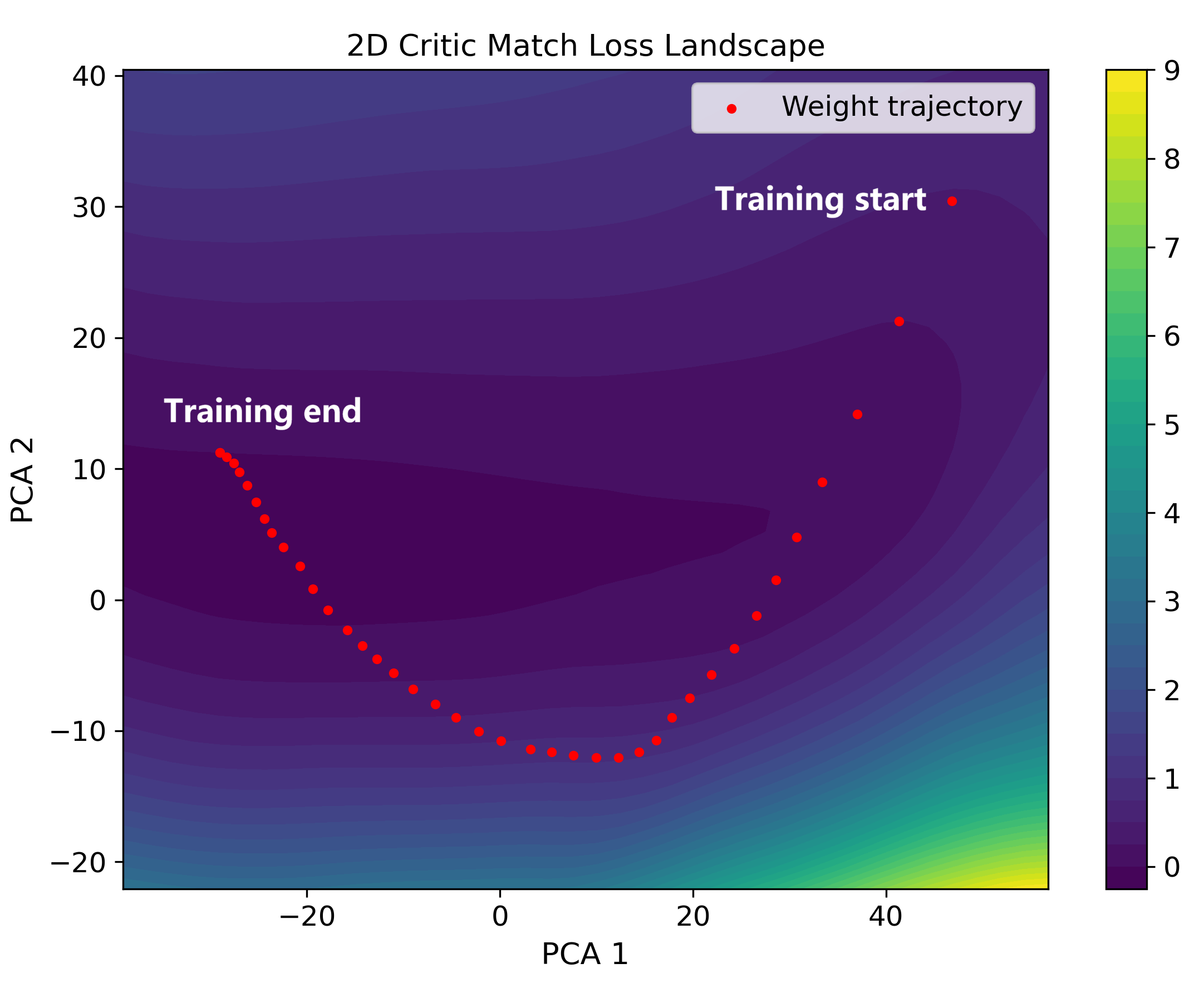}
        \label{fig:SAC_2D loss}
}
    \caption{Final 3-D and 2-D loss landscape of convergent spacecraft attitude SAC control}
    \label{fig:SAC_spc_loss landscap_conv}
\end{figure}

Using the method in \autoref {subsection: SAC loss landscape method}, the loss landscape under the final policy is generated for the SAC control process in \autoref{section: SAC attitude control setting}. To illustrate the result, only one critic network is selected to show the loss landscape during training. The weight is saved every 5000 steps during training. Data with batch size of 64 is generated as the final rollout after training is finished.  Following this setting, the loss landscape of SAC algrotihm is shown in \autoref{fig:SAC_spc_loss landscap_conv}.


From \autoref{fig:SAC_3D loss}, we can see that, the 3D loss landscape of SAC exhibits a broad, smooth basin with gentle curvature. It means small perturbations of the critic parameters produce limited variation in the match loss. This structure indicates robust convergence and insensitivity to local noise, which is benefited from the batch learning and entropy regularization technique that SAC employs. \autoref{fig:SAC_2D loss} illustrates the optimization path of the critic parameters over training. The trajectory lies in a continuous valley. The earlier iterations span in a wider region, while the later path converges into a narrow and stable zone. This evolution is consistent with the critic loss curve in \autoref{fig: SAC training curve}, which also shows steep loss reduction in the early stage and stabilization near the bottom of the basin.

\subsection{Loss landscape comparison between convergent SAC control and divergent ADHDP control}


\begin{figure}[htbp]
    \centering
    \subfigure[3-D loss landscape]{
        \centering
        \includegraphics[width=0.5\textwidth]{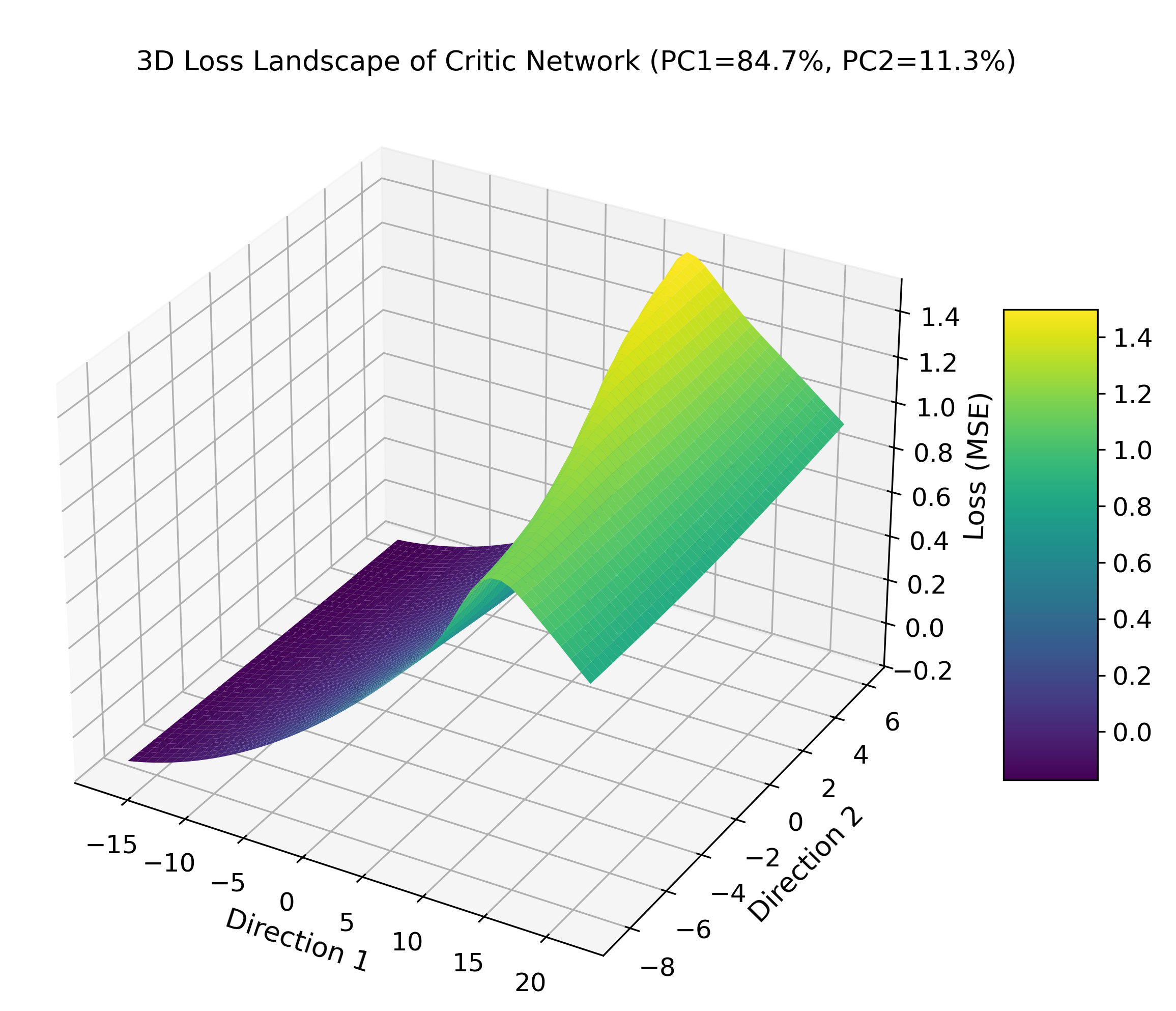}
        \label{fig:spc_3D loss}
}
    \hspace{0.02\columnwidth}
    \subfigure[2-D loss landscape]{
        \centering
        \includegraphics[width=0.5\textwidth]{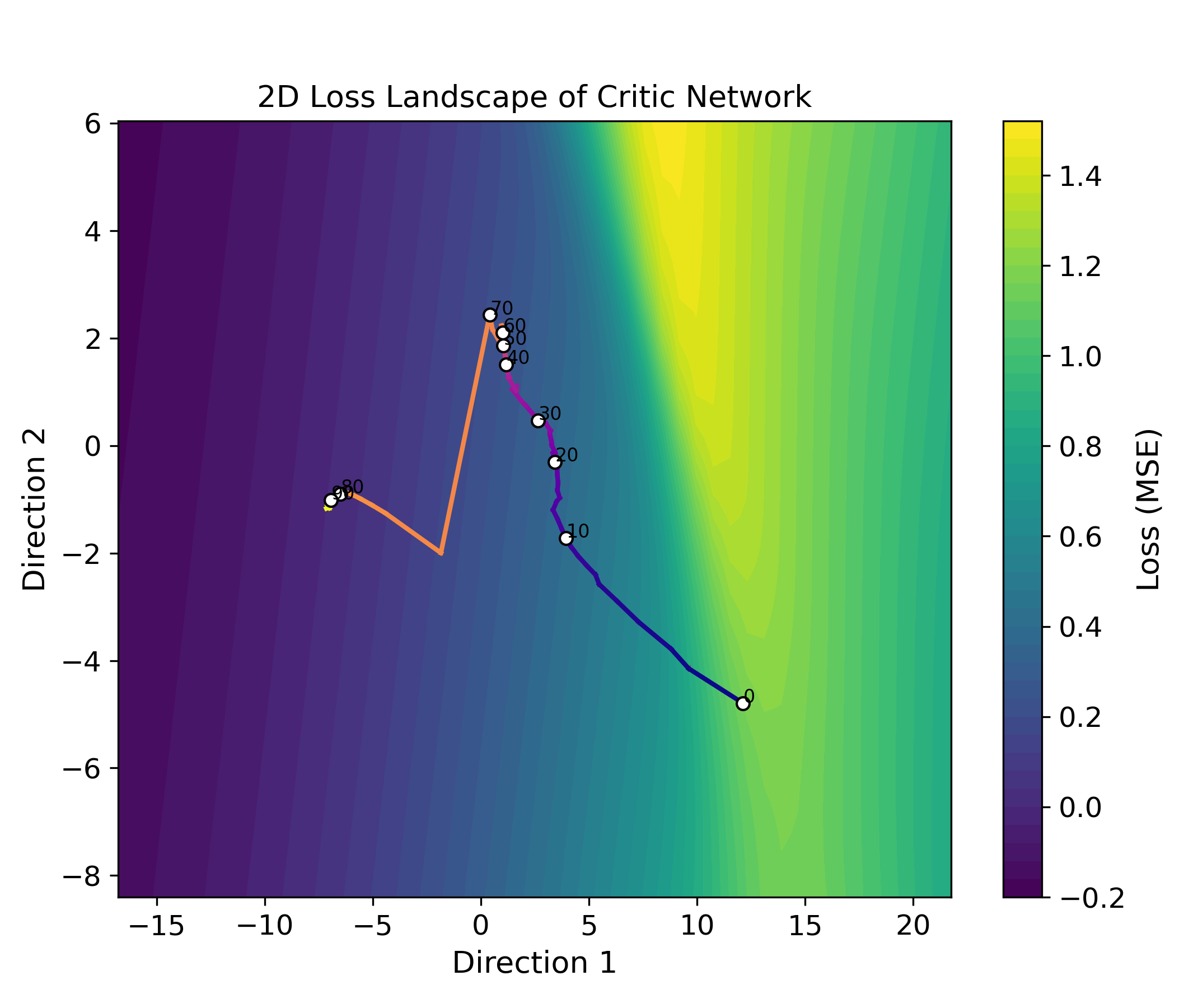}
        \label{fig:spc_2D loss}
}
    \caption{3-D and 2-D loss landscape of spacecraft attitude ADHDP control}
    \label{fig:spc_loss landscape}
\end{figure}

Using the same critic match loss landscape construction framework, \autoref{fig:spc_loss landscape} illustrates the 3-D loss landscape and the 2-D loss path resulting from applying the ADHDP algorithm to the spacecraft attitude system using with quaternion expression. The control results corresponding to the ADHDP critic loss landscape are shown in Appendix \ref{app:Appendix ADHDP control results}.

\autoref{tab:SAC_ADHDP_landscape_metrics_spc} shows the quantitative indices values of the landscapes of SAC and ADHDP algorithm on the spacecraft attitude control problem. $\rho$ value for calculating sharpness is selected as 0.25. $\epsilon$ value for calculating basin area is selected as 0.1. 

\begin{table}[t]
\centering
\caption{Quantitative loss landscape comparison between SAC and ADHDP on attitude control task.}
\label{tab:SAC_ADHDP_landscape_metrics_spc}
\begin{tabular}{lcc}
\hline
\textbf{Metric} & \textbf{SAC} & \textbf{ADHDP} \\
\hline
Sharpness ($\mathrm{Sharp}_{\epsilon}$) 
& 0.7255963360245651 
& 0.10840051628952453 \\

Basin ($A_{\rho}$) 
& 1252.2596440947382 
& Non-existent \\

Local anisotropy ($\log \kappa$) 
& 3.718889422035786 
& 10.39062500455904 \\
\hline
\end{tabular}
\end{table}

From the comparison between SAC's critic match loss landscape and ADHDP's landscape, there are three aspects that indicate obvious differences in the landscape of SAC algorithm and ADHDP algorithm.

The first aspect is the PCA range and the range of the optimization trajectory. For the ADHDP landscape, the small changes in the PCA direction indicate that the main direction of weight changes is relatively limited. It oscillates within a narrow parameter space. This often suggests limited exploration or oscillation within a restricted parameter manifold. In contrast, for the landscape of SAC algorithms, it has larger spans of weight updates in the main direction. The stronger contributions from the PCA principal components indicate that the network has explored a wider range of weight spaces and ultimately reached a low-loss region. Rather than indicating successful control solely  through span magnitude, this difference highlights distinct optimization behaviors under online and replay-based training mechanisms.

The second aspect is the curvature and structure of the loss landscape. The ADHDP loss landscape exhibits sharp curvature and irregular contour shapes. This observation is consistent with the quantitative metrics in ~\autoref{tab:SAC_ADHDP_landscape_metrics_spc}. The relatively high local anisotropy value $\log \kappa = 10.39$ indicates strong directional imbalance in curvature, meaning that gradients can change abruptly depending on direction. Moreover, the absence of a well-defined basin suggests that no stable low-loss region is formed. Together, these characteristics reflect unstable gradient behavior and sensitivity to perturbations during training. By contrast, the SAC loss landscape shows a clear concave structure with a single valley. Within this valley, the surface is relatively flat and wide. From ~\autoref{tab:SAC_ADHDP_landscape_metrics_spc}, this valley has a large basin area with $A_{\rho} = 1252.26$ and a much lower anisotropy value with $\log \kappa = 3.72$, indicating more balanced curvature across directions. Although the sharpness value of SAC is higher than that of ADHDP, it is associated with a structured descent toward a well-defined basin rather than irregular local fluctuations. Such a geometry is associated with a more structured optimization region under the frozen-target objective, where parameter updates are less sensitive to directional imbalance compared to the ADHDP case.

The third aspect is the range of the loss. In the loss landscape, the scale of variation between the highest and lowest loss for the ADHDP algorithm is considerably smaller than that for the SAC algorithm. It means that, during training of the ADHDP algorithm, it explores within a limited zone, of which the loss is not so much different. However, for the SAC algorithm, the large changes in loss indicate that the optimization goes through a more complete procedure, which reflects a more extended scan of the solution subspace. It has to be noted that, the presence of negative values in the normalized loss surface of ADHDP indicates that the final critic parameters are not located at a local minimum within the scanned subspace. Consequently, the loss landscape does not exhibit a closed basin structure around the final point.

These results show that the proposed critic match loss landscape visualization method remains applicable in the off-policy reinforcement learning setting. The smooth basin revealed in the SAC critic match loss landscape corresponds to a smooth optimization region and a convergent critic approximation, which is consistent with SAC’s replay-based updates and target smoothing technique. In contrast, the sharper geometry observed in the ADHDP landscape shows the sensitivity of online adaptive learning to data noise and unstable target updates. The comparison between the two algorithms shows that the proposed visualization method can interpret the critic learning stability for actor–critic reinforcement learning algorithms. It also demonstrates that the method remains effective across different training paradigms.

\subsection{Temporal evolution of critic optimization of divergent SAC control}

The geometric characteristics of critic optimization of the convergent SAC case under the proposed visualization framework is illustrated in \autoref{subsection:convergent_SAC}. To further examine how the critic match loss landscape behaves under different training outcomes within the same algorithmic structure, we next consider two divergent SAC cases obtained under altered training conditions. To investigate how the geometry develops over time, critic match loss landscapes are constructed at multiple training stages. The temporal evolution of the projected critic parameter trajectory is then examined to reveal changes in optimization behavior as divergence emerges.

For the divergent-control experiments, the learning rate was increased to $2\times10^{-1}$, the replay buffer size was reduced to $5\times10^{2}$, and the batch size was set to $16$. A large discount factor $\gamma = 0.999$ and a fast target update coefficient $\tau = 0.999$ were used. Entropy regularization was fixed to $\alpha = 10^{-3}$ without automatic tuning. Actor and critic networks were updated at every step with 10 gradient iterations. This configuration produced divergent attitude responses.

\subsubsection{Divergent SAC with critic parameter divergence}
\label{subsubsection:divSACwithdivcritic}

We now examine a divergent SAC realization in which critic instability develops with pronounced parameter divergence. While the algorithmic structure remains unchanged, this training outcome exhibits fundamentally different optimization behavior within the critic network.

\begin{figure*}[htbp]
    \centering
    \subfigure[SAC control result]{
        \centering
        \includegraphics[width=0.6\textwidth]{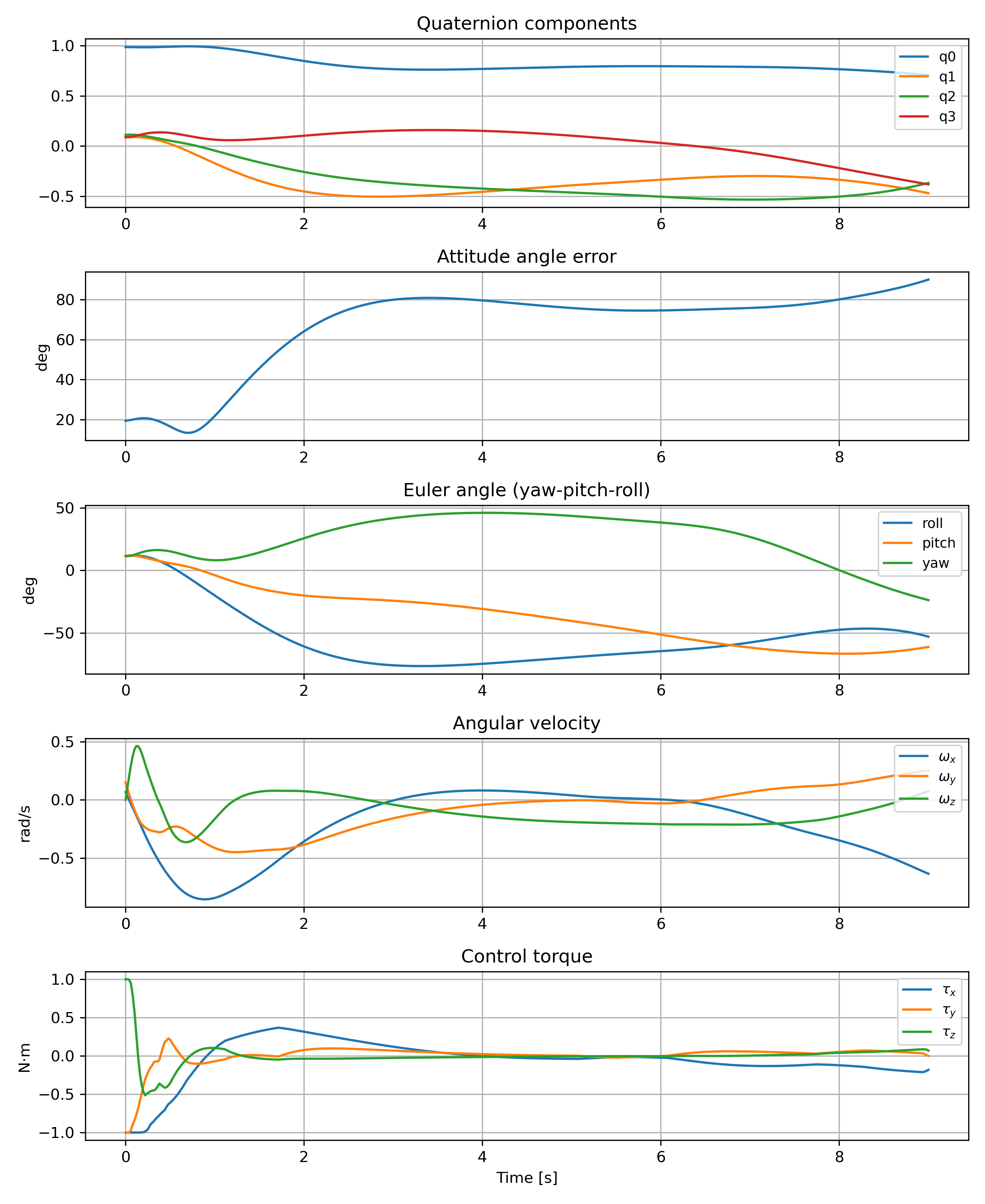}
        \label{fig:divSACdivcritic sc system performance}
}
    \hspace{0.02\columnwidth}
    \subfigure[SAC training curve]{
        \centering
        \includegraphics[width=0.6\textwidth]{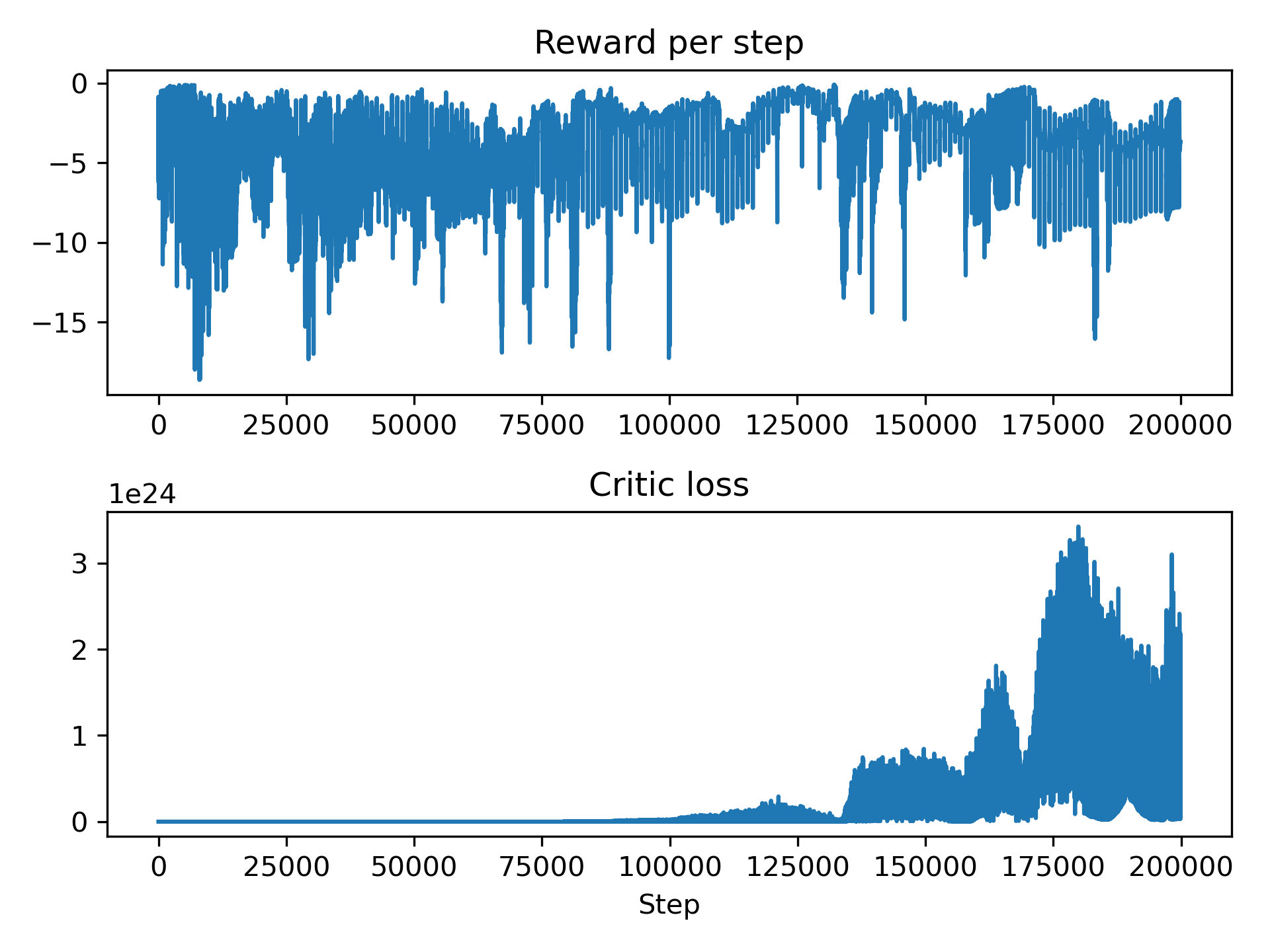}
        \label{fig: divSACdivcritic sc training curve}
}

    \caption{ Divergent control results for spacecraft attitude system under SAC with divergent critic}
    \label{fig:divSACdivcritic sc system and algorithm performance}
\end{figure*}

\begin{figure}[htbp]
    \centering

    \subfigure[critic weight with training steps]{
        \includegraphics[width=0.48\textwidth]{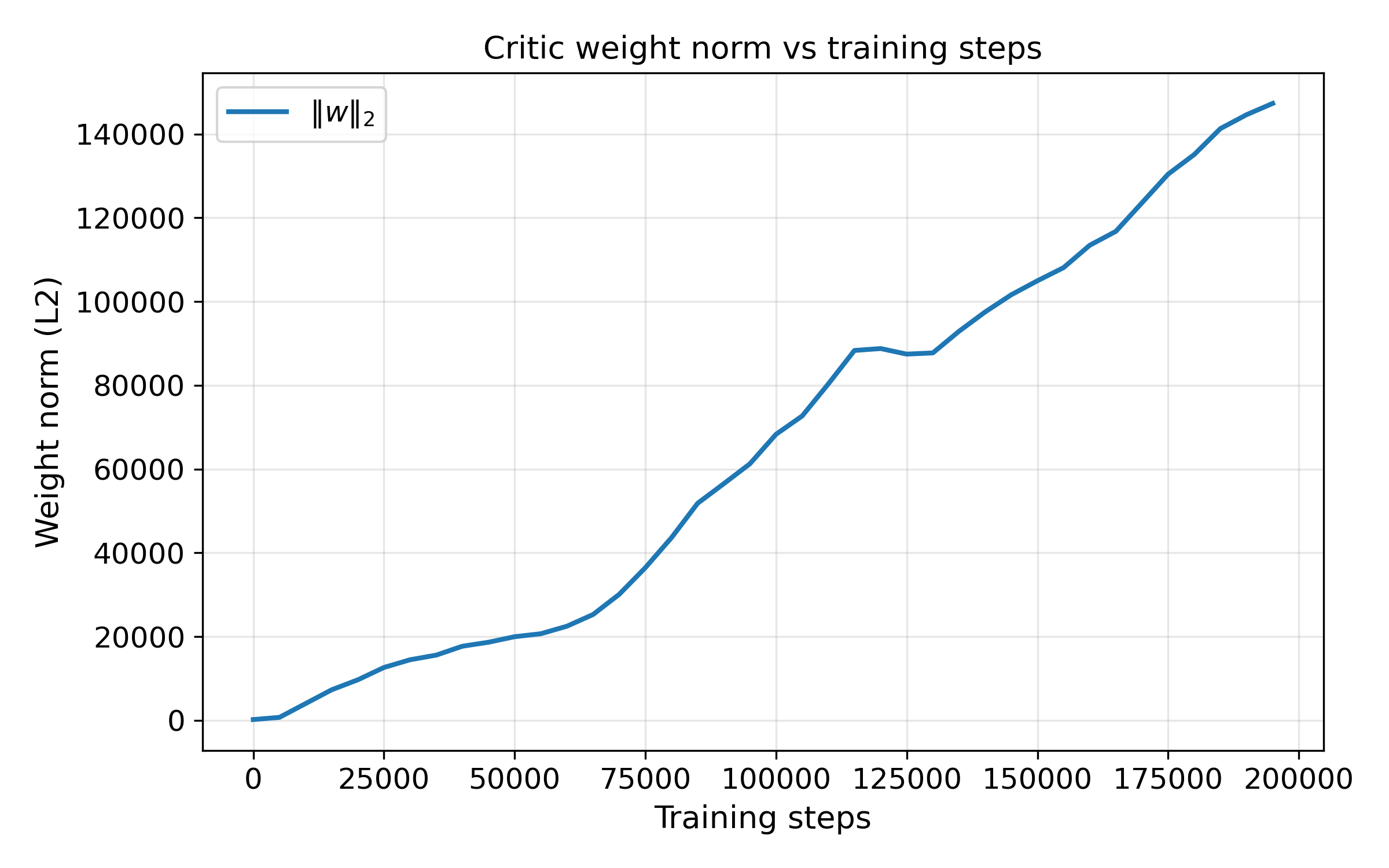}
        \label{fig:divSACdivcritic_criticweights}
    }%
    \hfill
    \subfigure[actor weight with training steps]{
        \includegraphics[width=0.48\textwidth]{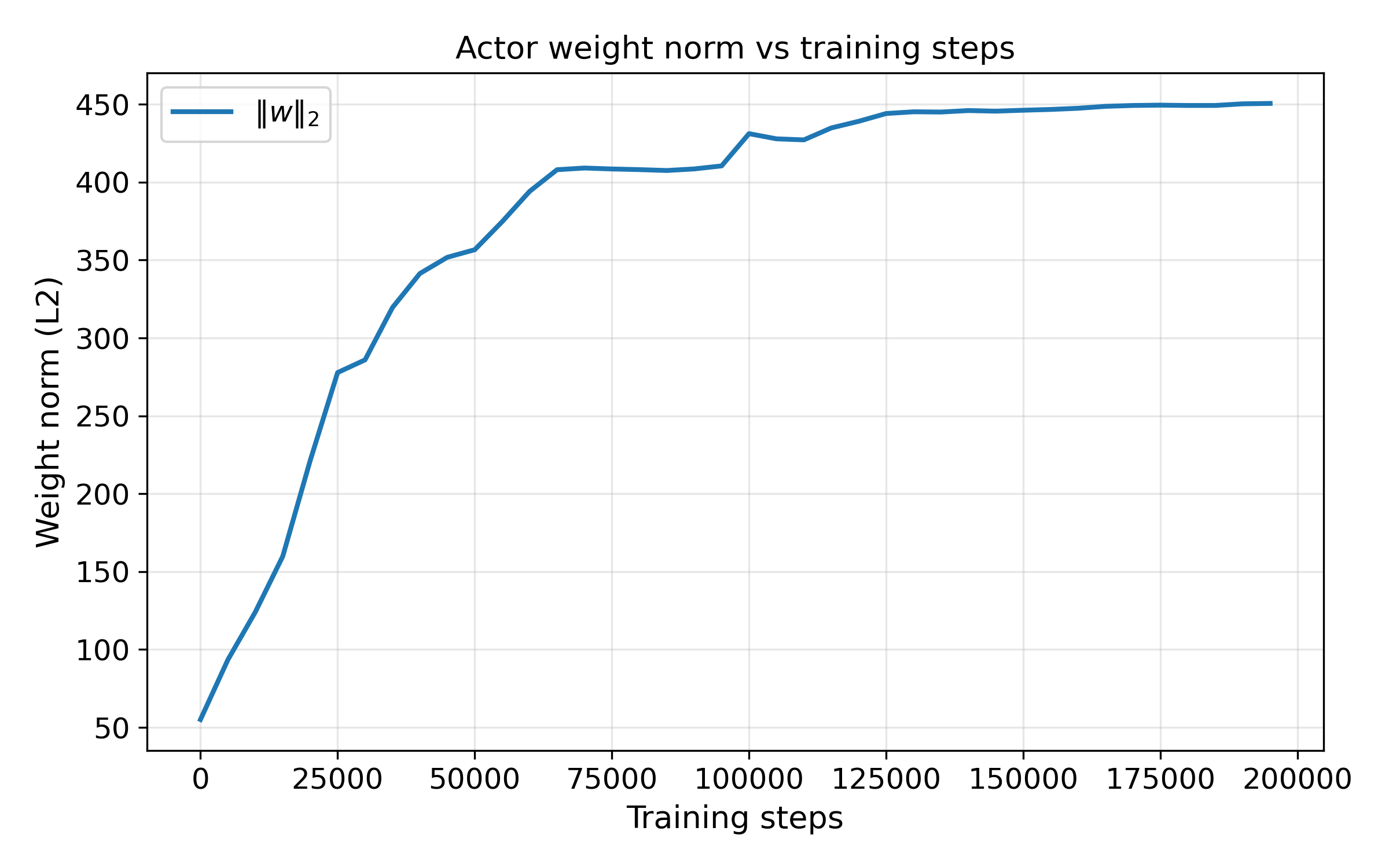}
        \label{fig:divSACdivcritic_actorweights}
    }

    \caption{Critic and actor weight with training steps for SAC with divergent critic}
    \label{fig:divSACdivcritic_critic_actor_weightwithsteps}
\end{figure}

\autoref{fig:divSACdivcritic sc system and algorithm performance} shows the final rollout trajectory of the attitude system and the training curves. From the closed-loop perspective, the final rollout demonstrates clear failure of attitude stabilization. The attitude angle error increases over time rather than converging toward zero, and the angular velocity components do not decay. The control torque signals do not form a stabilizing feedback pattern, indicating that the learned policy fails to stablize the spacecraft dynamics. The scalar training curves further reveal instability. The critic loss grows to extremely large magnitudes during later stages of training, and the reward does not exhibit sustained improvement. \autoref{fig:divSACdivcritic_critic_actor_weightwithsteps} illustrates the critic weight and actor weight update with training steps. Consistently, the critic weight norm increases almost monotonically throughout training and reaches a significantly larger scale than in the convergent case. In contrast, the actor weight norm grows more moderately and eventually stabilizes. This divergence of critic parameter magnitude suggests directional runaway rather than bounded oscillation. However, scalar curves alone do not reveal along which parameter directions this instability develops, nor how it relates to the geometry of the optimization landscape.

\begin{figure}[htbp]
    \centering
    \subfigure[3-D loss landscape for divergent SAC control with divergent critic]{
        \centering
        \includegraphics[width=0.5\textwidth]{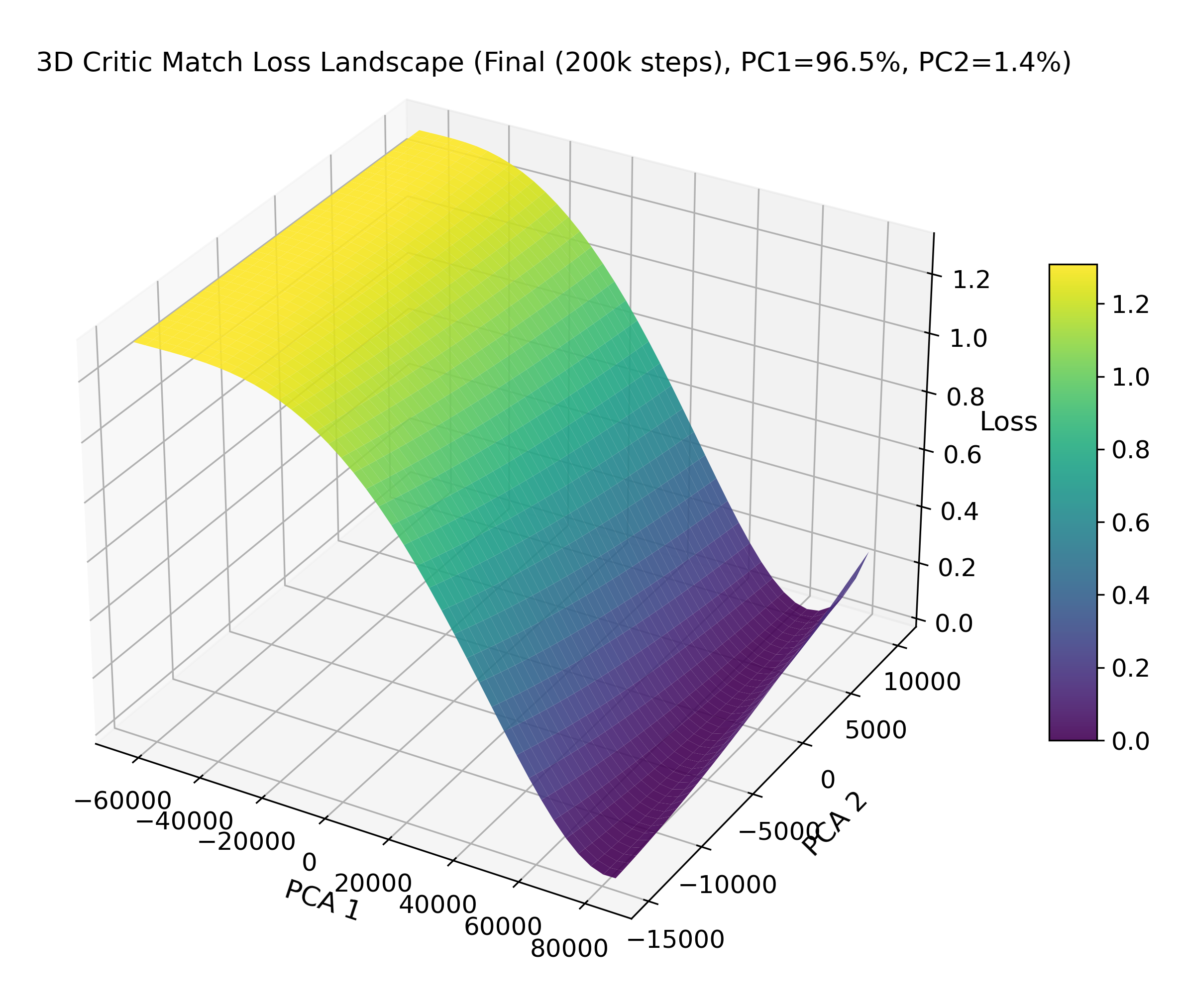}
        \label{fig:SAC_3D loss_divSACdivcritic}
    }
    \hspace{0.02\columnwidth}
    \subfigure[2-D loss curve for divergent SAC control with divergent critic]{
        \centering
        \includegraphics[width=0.5\textwidth]{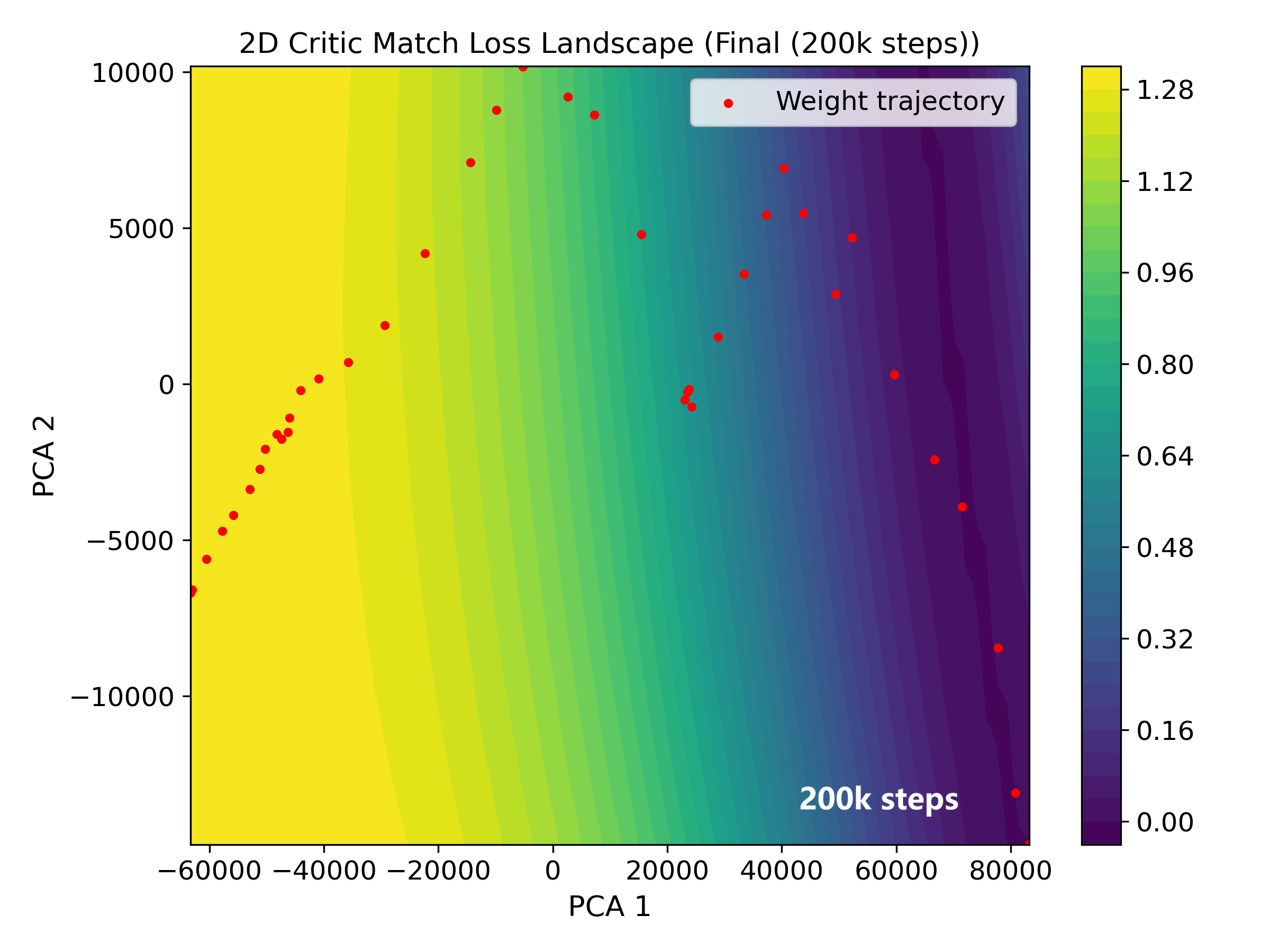}
        \label{fig:SAC_2D loss_divSACdivcritic}
}
    \caption{3-D and 2-D loss landscape of divergent spacecraft control under SAC with divergent critic}
    \label{fig:SAC_spc_loss landscap_divSACdivcritic}
\end{figure}

The final 3D critic match loss landscape in \autoref{fig:SAC_3D loss_divSACdivcritic} provides essential geometric insight. Under the frozen-target approximation, the surface does not exhibit a well-defined closed basin structure. Instead, it resembles a strongly anisotropic slope or ridge extending along the dominant principal direction. The PCA variance analysis indicates that the first principal component accounts for $96.5\%$ of the total variance, while the second component contributes only marginally. This extreme dominance of PC1 implies that critic parameter evolution is effectively confined to a nearly one-dimensional subspace. In such a geometrically degenerate setting, parameter updates are driven primarily along a single direction. If this dominant direction does not correspond to a stable descent toward a basin, persistent amplification along it can lead to uncontrolled parameter growth. Importantly, this near one-dimensional update behavior is not directly observable from the scalar weight-norm curve. It becomes evident only through the PCA structure of the loss landscape.

The corresponding 2D projection in \autoref{fig:SAC_2D loss_divSACdivcritic} further clarifies the parameter evolution. The loss contours lack closed low-loss regions indicative of a stable basin. The projected critic trajectory spans a large distance primarily along the PC1 direction, with limited variation along PC2. Rather than exhibiting discrete regime transitions between separated clusters, the trajectory displays sustained directional drift. This behavior is consistent with the PCA variance structure and explains the monotonic increase in critic weight norm that the optimization process is not exploring multiple regions of the landscape, but instead being progressively driven along a dominant and geometrically sensitive axis.

\begin{figure}[htbp]
\centering
\subfigure[3-D landscape at step 20000 ]{
    \includegraphics[width=0.48\textwidth]{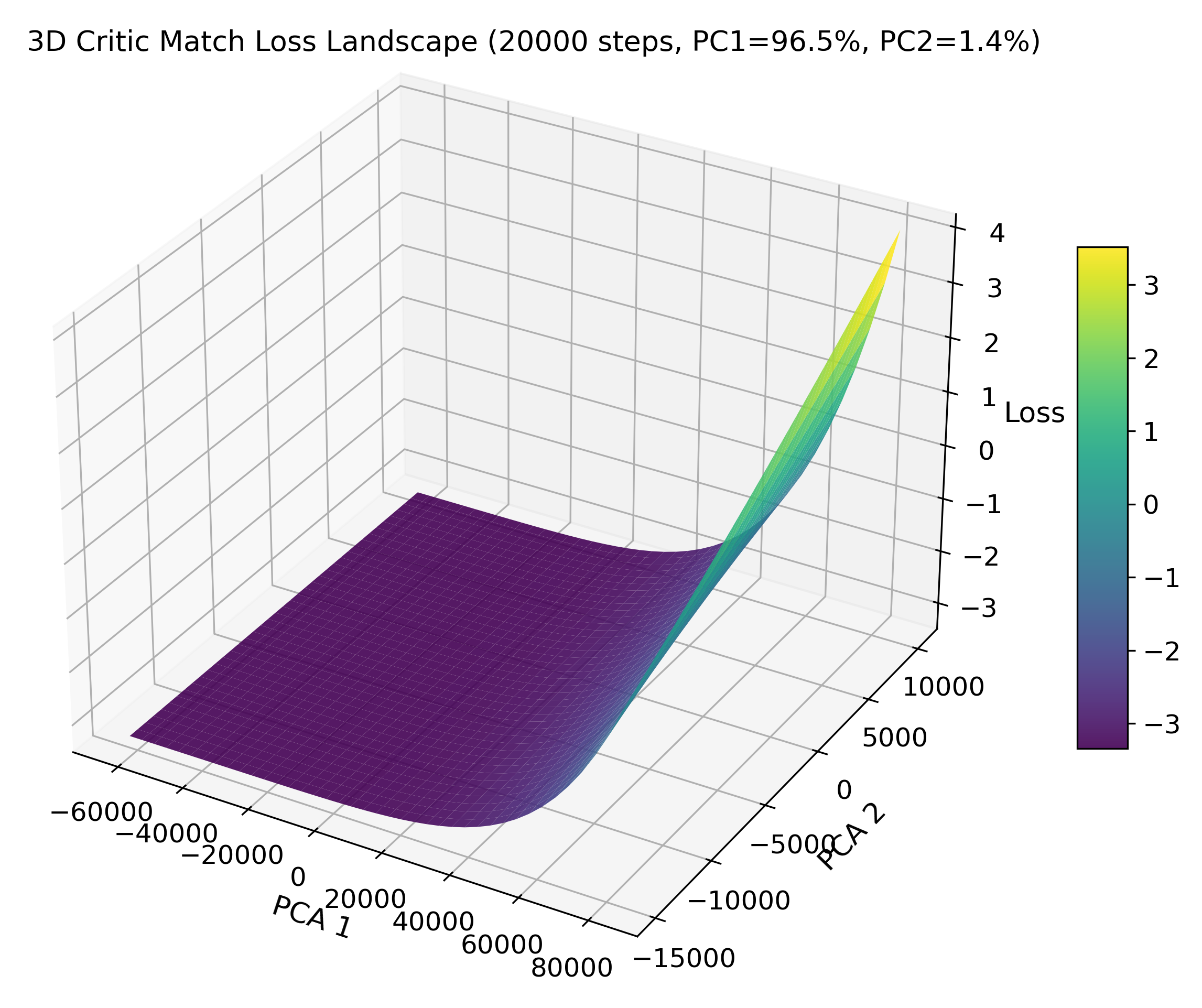}
    \label{fig:SAC_3d_step20000_divSACdivcritic}
}
\hfill
\subfigure[2-D landscape at step 20000]{
    \includegraphics[width=0.48\textwidth]{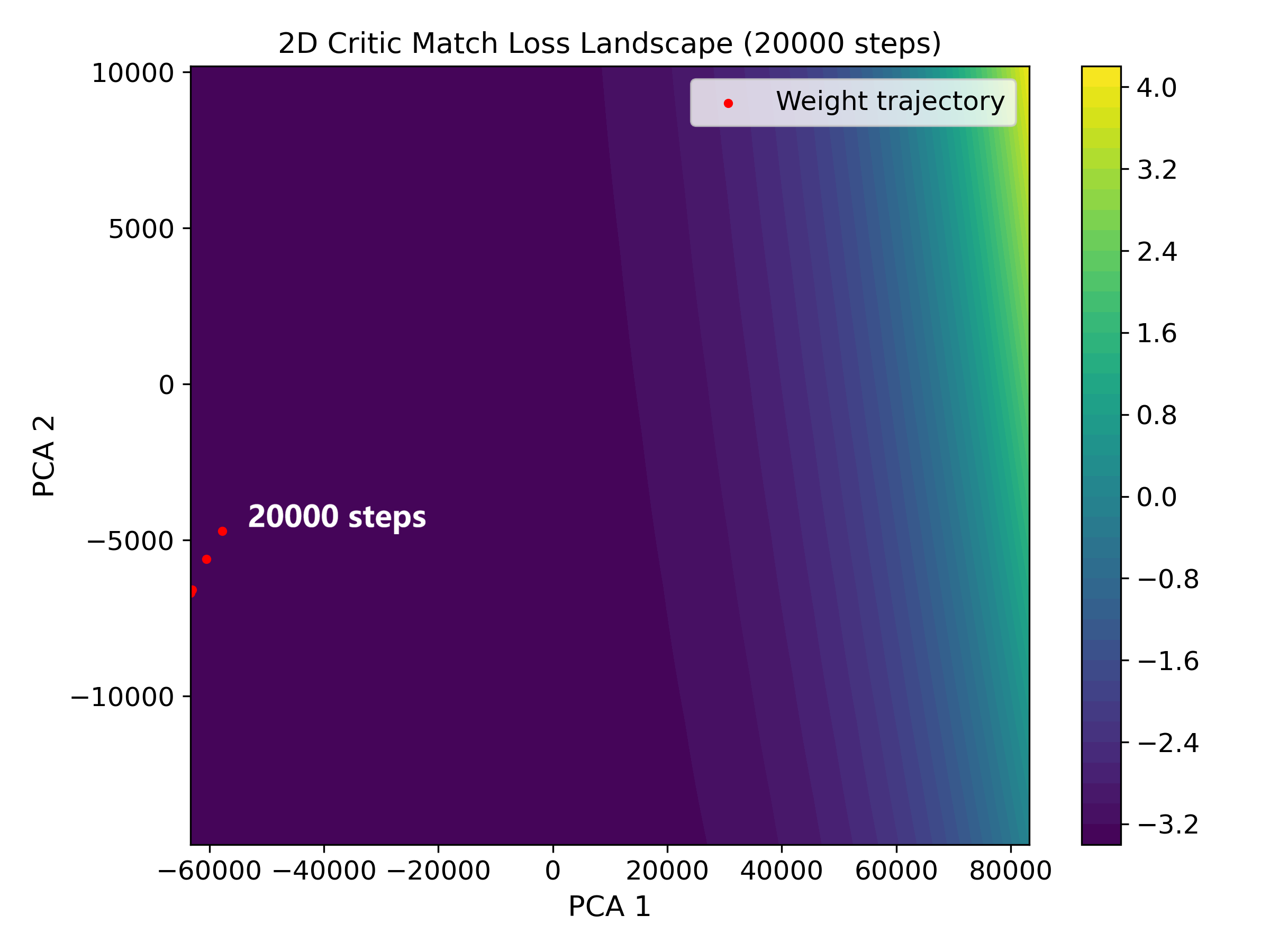}
    \label{fig:SAC_2d_step20000_divSACdivcritic}%
}

\subfigure[3-D landscape at step 50000]{
    \includegraphics[width=0.48\textwidth]{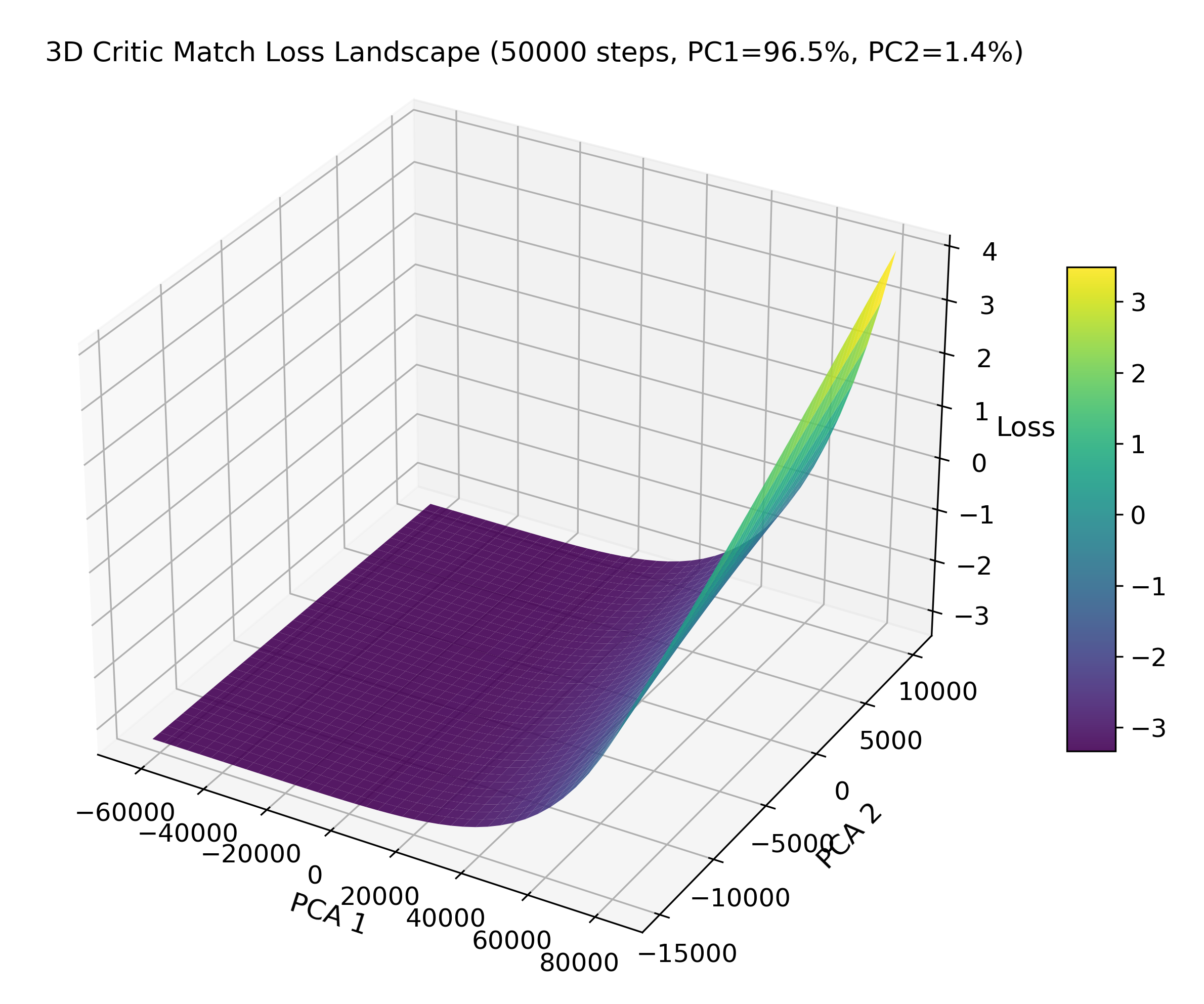}
    \label{fig:SAC_3d_step50000_divSACdivcritic}
}
\hfill
\subfigure[2-D landscape at step 50000]{
    \includegraphics[width=0.48\textwidth]{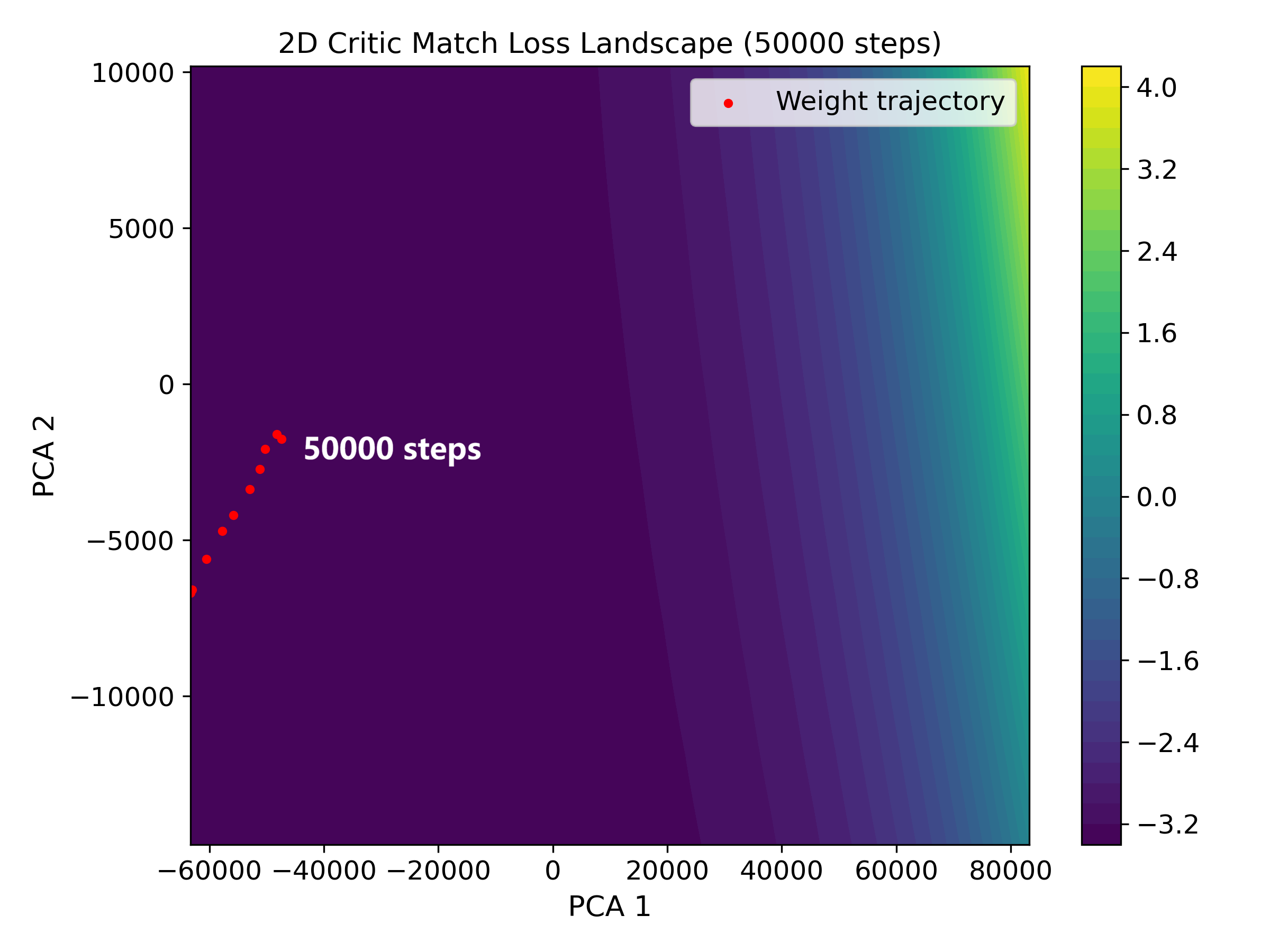}
    \label{fig:SAC_2d_step50000_divSACdivcritic}
}

\subfigure[3-D landscape at step 100000]{
    \includegraphics[width=0.48\textwidth]{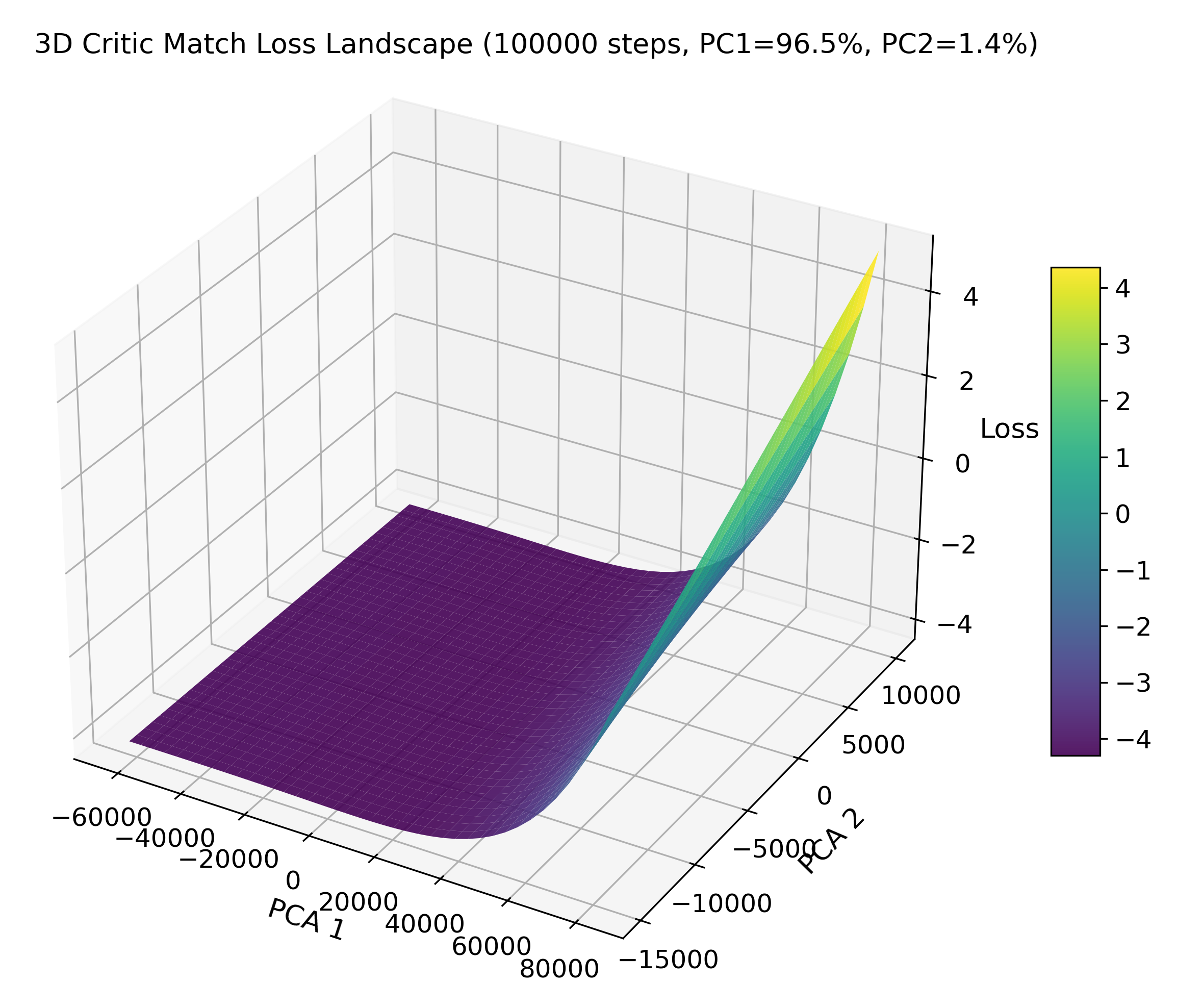}
    \label{fig:SAC_3d_step100000_divSACdivcritic}
}
\hfill
\subfigure[2-D landscape at step 100000]{
    \includegraphics[width=0.48\textwidth]{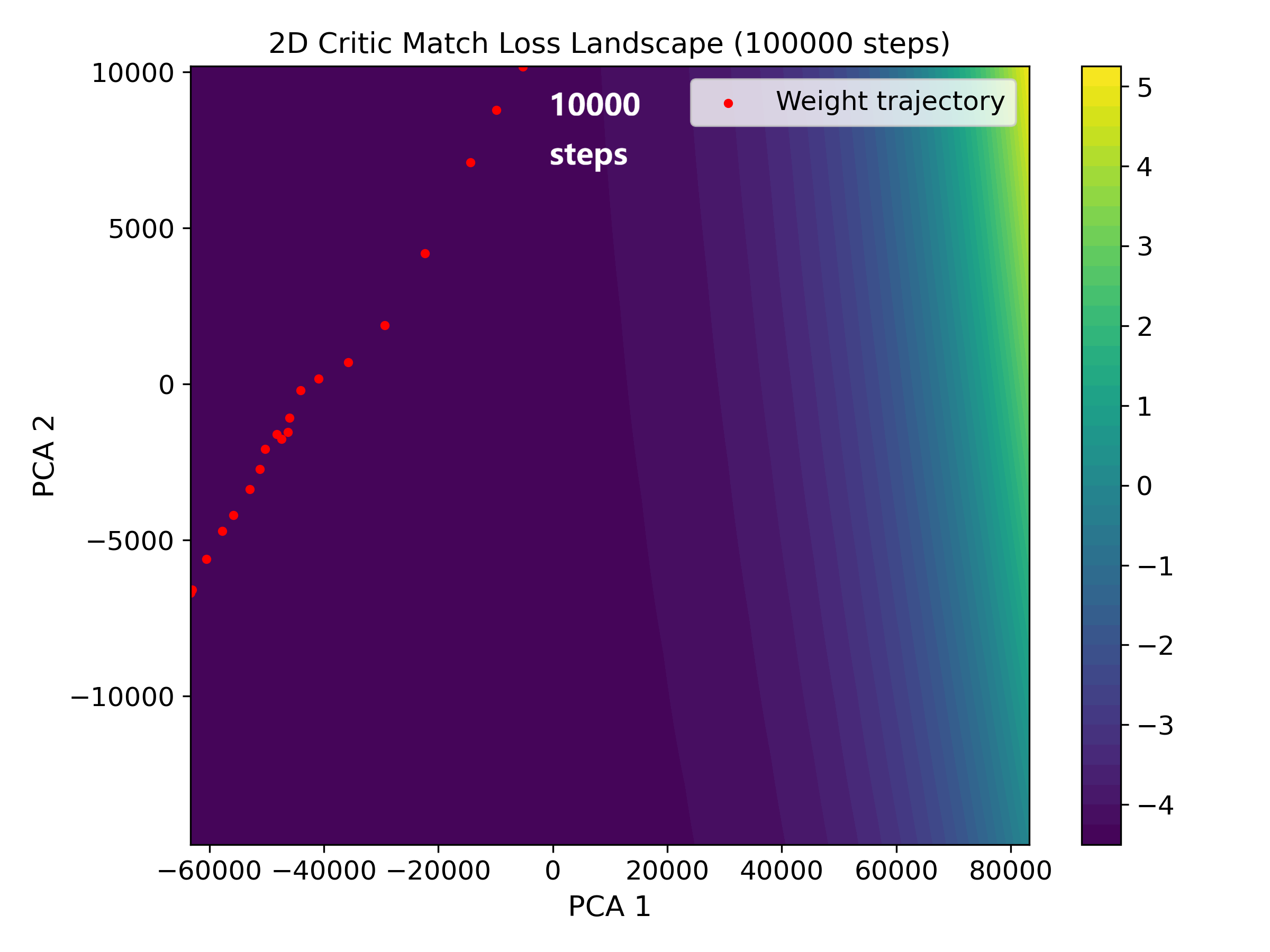}
    \label{fig:SAC_2d_step100000_divSACdivcritic}
}

\caption{Snapshots during training of 3-D and 2-D loss landscape of divergent spacecraft attitude control under SAC with divergent critic}
\label{fig:SAC_spc_loss landscap__divSACdivcritic_snapshots}

\end{figure}

To interpret the temporal snapshots in \autoref{fig:SAC_spc_loss landscap__divSACdivcritic_snapshots}, it is important to recall that the PCA plane is constructed using critic weights collected over the entire training process. All snapshots are therefore projected onto a common global coordinate frame, ensuring comparability of trajectory evolution. The reported variance ratio remains constant across snapshots.Because this global PCA basis is influenced by late-stage parameter excursions, early-stage grid points may include regions far from the contemporaneous parameter neighborhood. Consequently, the temporal interpretation should not rely on absolute colorbar ranges. Instead, emphasis is placed on trajectory alignment and geometric structure. 
\autoref{fig:SAC_spc_loss landscap__divSACdivcritic_snapshots} demonstrates temporal snapshots of the critic match loss landscape which are generated at 20k, 50k, 100k, and 200k steps. Across snapshots, the critic trajectory continues to move along the same dominant principal direction, without forming a closed basin, suggesting sustained directional growth rather than stable convergence.

Overall, the final and temporal critic match loss landscapes of the divergent SAC control with divergent critic parameters demonstrate that when the critic match loss landscape exhibits strong anisotropy and effective dimensional collapse in parameter space, critic optimization may become unstable and lead to runaway parameter growth. In this case, the geometrically degenerate structure of the frozen-target landscape aligns with divergent critic parameters and closed-loop control failure. The critic match loss visualization therefore serves as a geometric diagnostic tool, revealing structural instability in the critic optimization process that is not directly visible from scalar training curves alone.

\subsubsection{Divergent SAC with critic parameter convergence}
\label{subsubsection:divSACwithconcritic}

\begin{figure*}[htbp]
    \centering
    \subfigure[SAC control result]{
        \centering
        \includegraphics[width=0.6\textwidth]{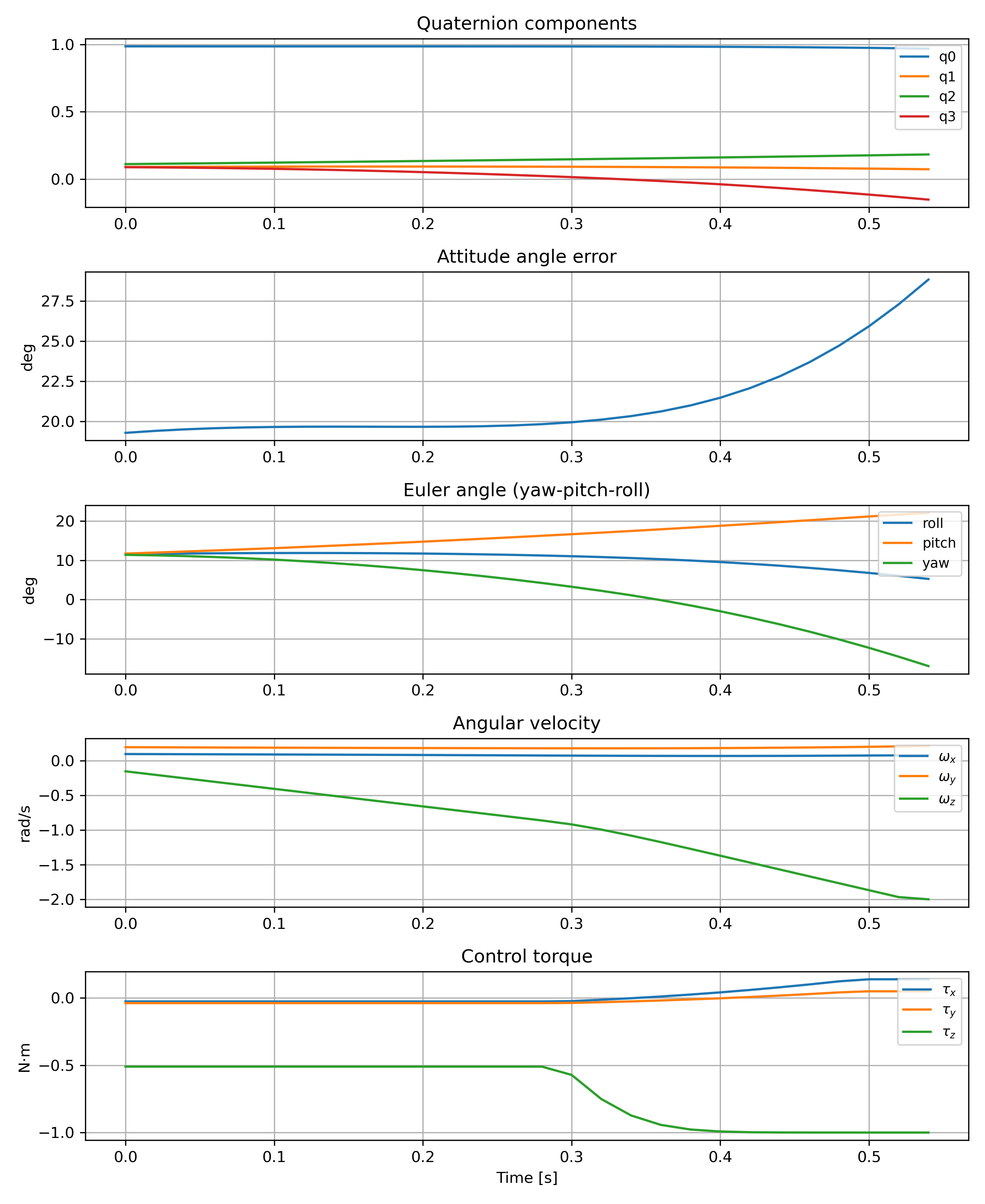}
        \label{fig: SAC divergent sc system performance}
}
    \hspace{0.02\columnwidth}
    \subfigure[SAC training curve]{
        \centering
        \includegraphics[width=0.6\textwidth]{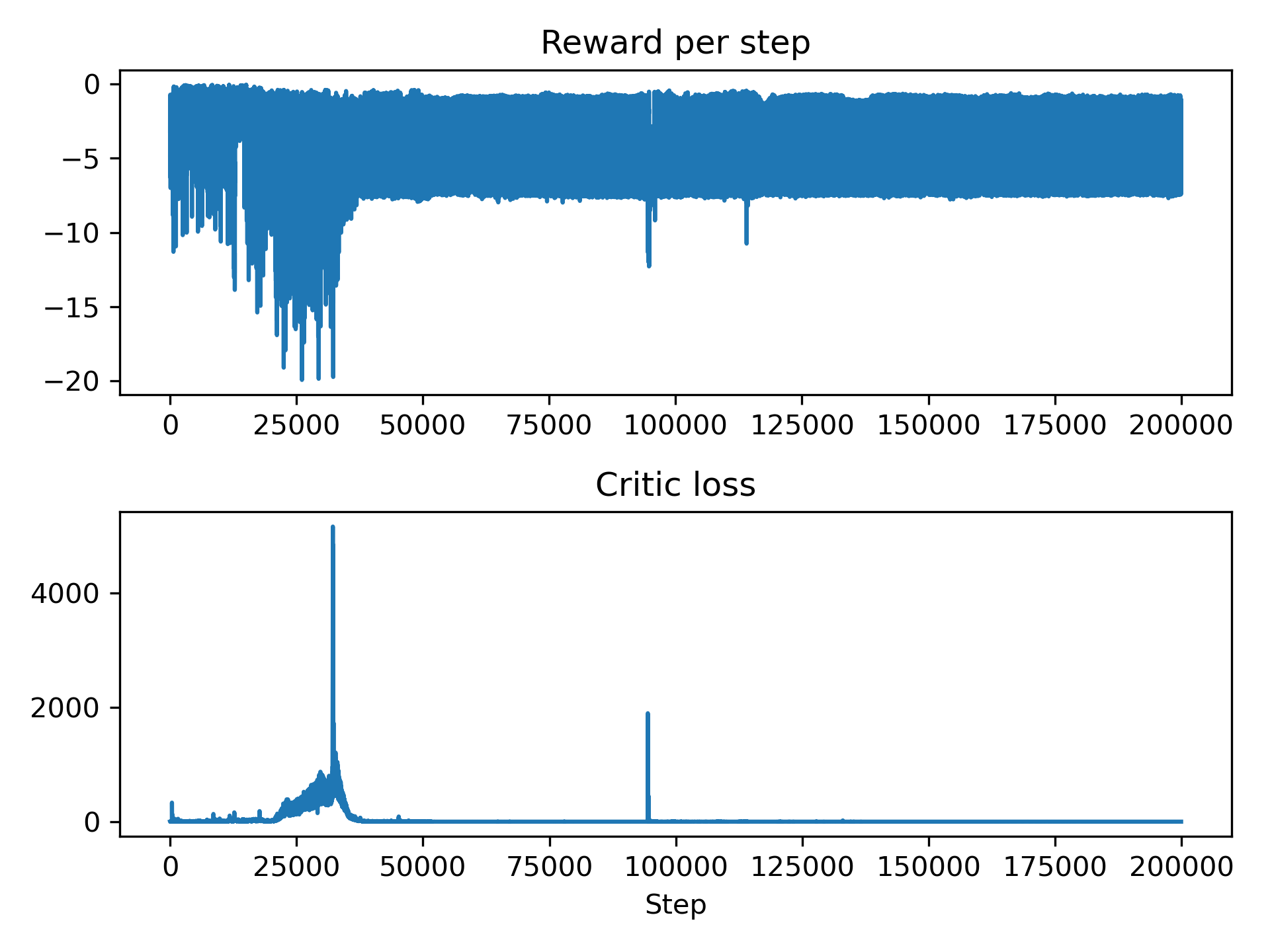}
        \label{fig: SAC divergent sc training curve}
}

    \caption{Divergent control results for spacecraft attitude system under SAC with convergent critic}
    \label{fig:SAC divergent sc system and algorithm performance}
\end{figure*}

In the divergent SAC case, as indicated by \autoref{fig: SAC divergent sc system performance}, the spacecraft attitude does not converge to the desired equilibrium. The final rollout shows that the attitude angle error increases over time rather than decreasing. The angular velocity components do not decay toward zero, and a clear drift is observed along the $\bm{\overline{z}}$ axis, or the $\bm{\overline{n_3}}$ axis as mentioned in \autoref{app:Appendix attitude dynamics}. The corresponding control torque remains close to its boundary for periods, indicating that the learned policy does not establish a stabilizing feedback law. Closed-loop control therefore fails.

The scalar training curves in \autoref {fig: SAC divergent sc training curve} provide additional evidence of instability. Although the critic loss eventually decreases to a relatively small value, several pronounced spikes occur during training. One large spike appears around 30k steps, and another significant excursion is observed near 100k steps. These spikes are substantially higher than surrounding values and reflect abrupt transitions in critic update dynamics rather than gradual convergence. The reward curve also remains persistently unfavorable, without a clear monotonic improvement trend.

\begin{figure}[htbp]
    \centering
    \subfigure[3-D loss landscape for divergent SAC control]{
        \centering
        \includegraphics[width=0.5\textwidth]{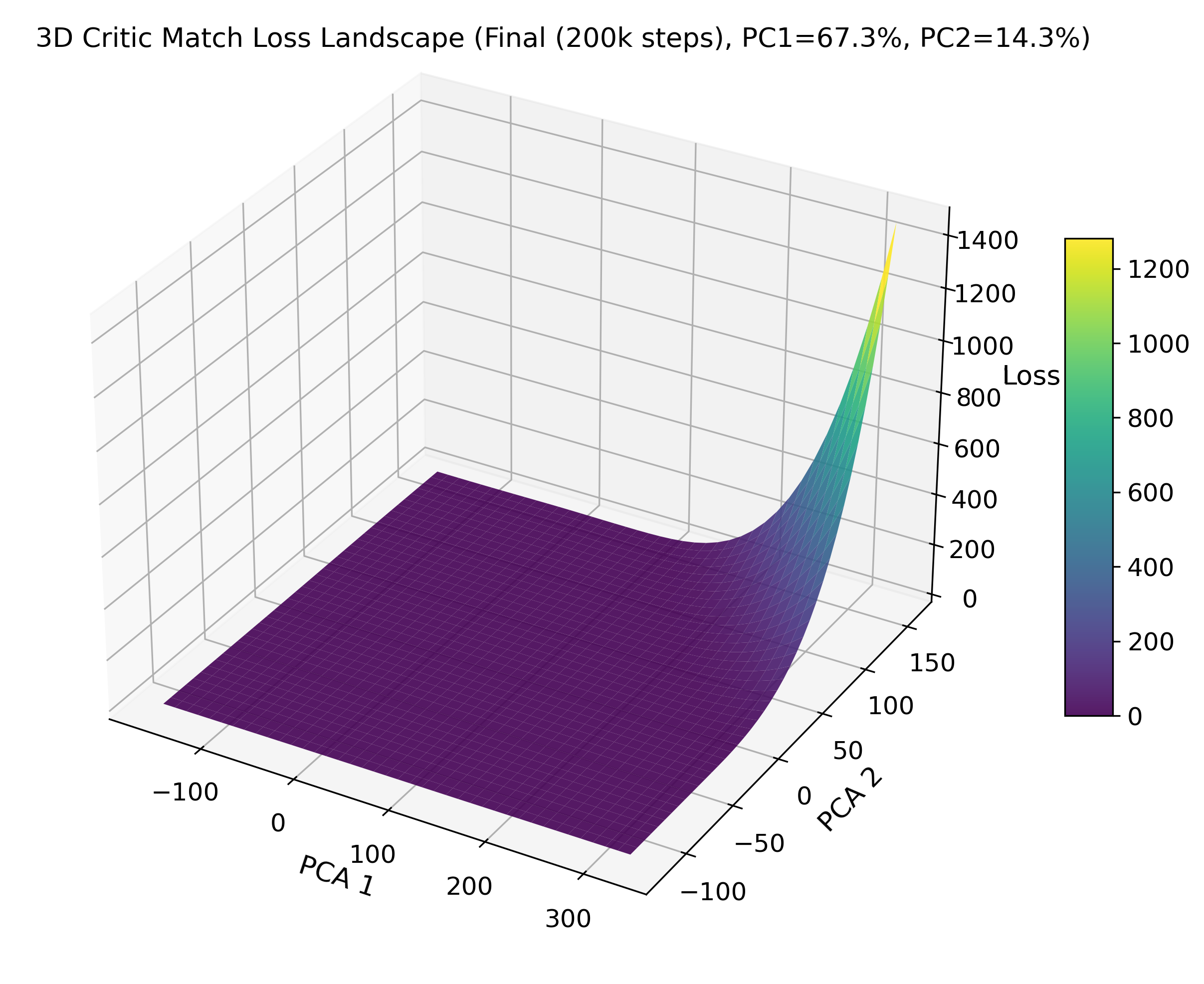}
        \label{fig:SAC_3D loss_div}
    }
    \hspace{0.02\columnwidth}
    \subfigure[2-D loss curve for divergent SAC control]{
        \centering
        \includegraphics[width=0.5\textwidth]{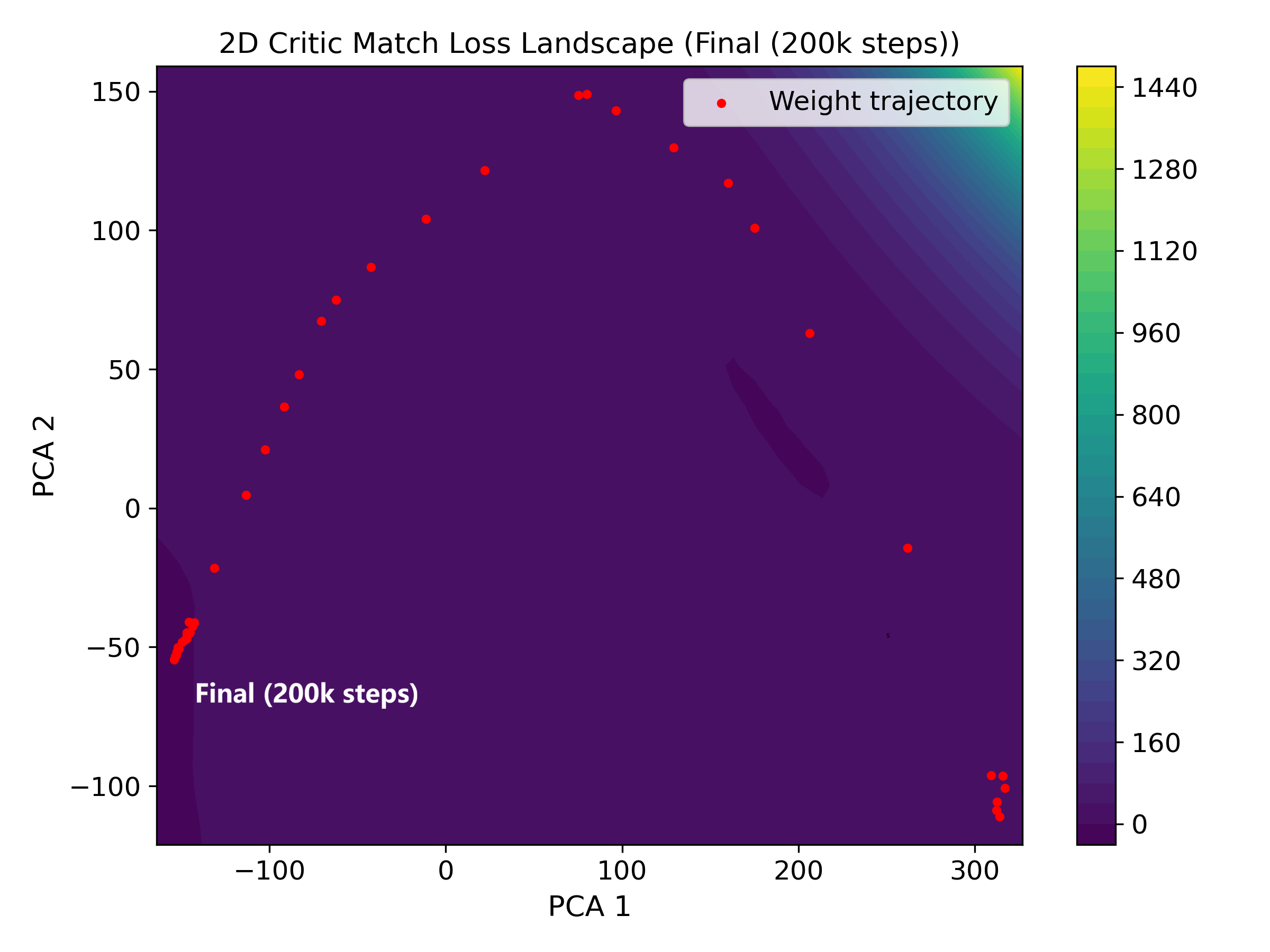}
        \label{fig:SAC_2D loss_div}
}
    \caption{3-D and 2-D loss landscape of divergent spacecraft attitude control under SAC with convergent critic}
    \label{fig:SAC_spc_loss landscap_div}
\end{figure}

To interpret these behaviors from the perspective of critic optimization, the critic match loss landscape is evaluated at the final stage, which corresponds to 200k steps, under a fixed replay batch and frozen target. As indicated by \autoref{fig:SAC_spc_loss landscap_div}, the final landscape exhibits broad smooth regions in the PCA plane together with steep high-loss walls in certain directions. The projected critic parameter trajectory does not form a single continuous curve. Instead, it appears separated into distinct clusters with large spatial gaps between them. The presence of separated clusters suggests that the critic underwent multiple discontinuous transitions during training rather than evolving smoothly toward a single stable region.

Importantly, the final landscape topology itself does not appear fundamentally different from convergent SAC cases. The failure therefore cannot be attributed to a rugged critic geometry. Instead, instability arises from discontinuities in the optimization path. The critic evolves across separated regions of the critic match loss landscape, leading to regime shifts that are not apparent from scalar loss values alone.

\begin{figure}[htbp]
\centering
\subfigure[3-D landscape at step 20000 ]{
    \includegraphics[width=0.48\textwidth]{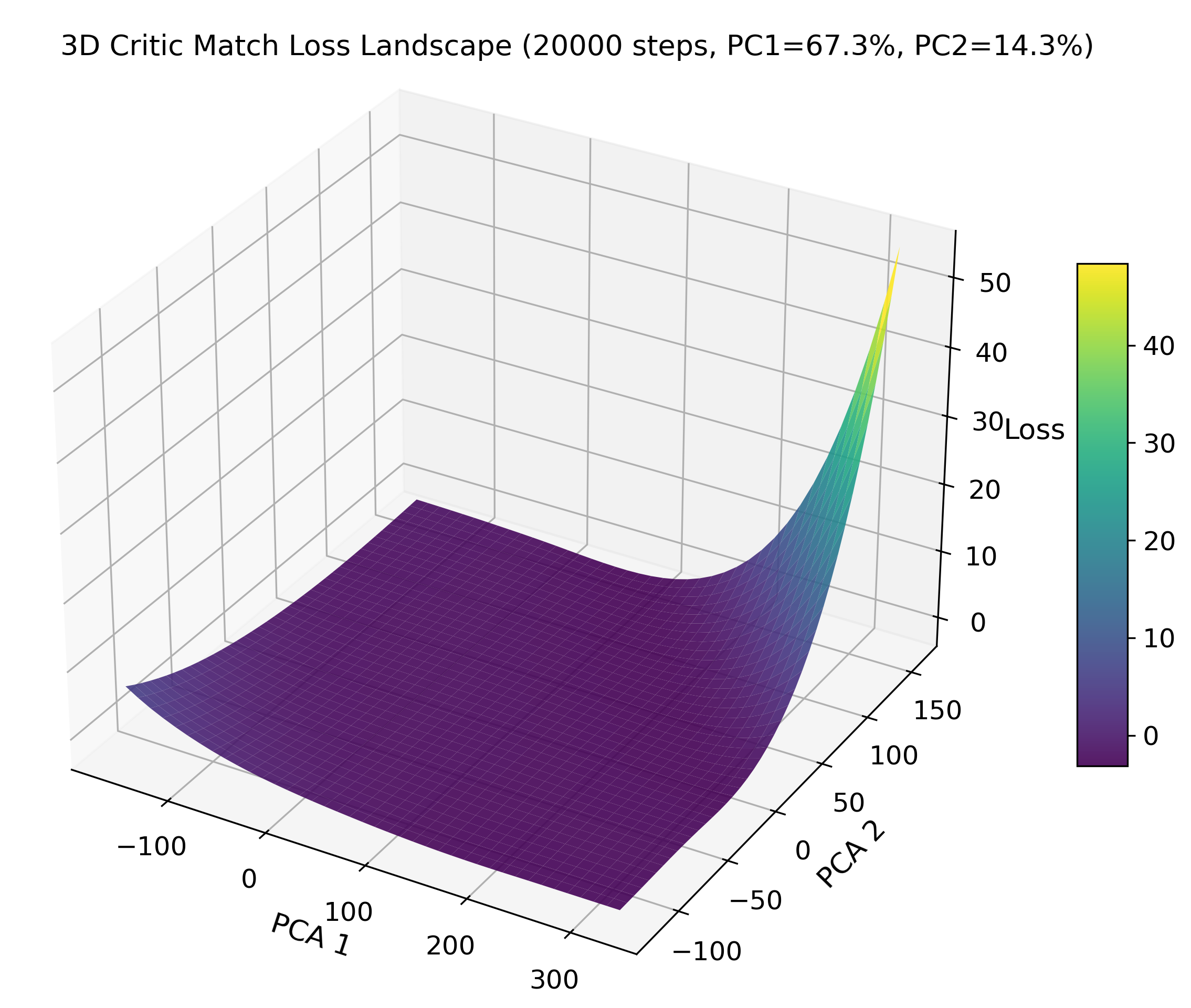}
    \label{fig:SAC_3d_step20000}
}
\hfill
\subfigure[2-D landscape at step 20000]{
    \includegraphics[width=0.48\textwidth]{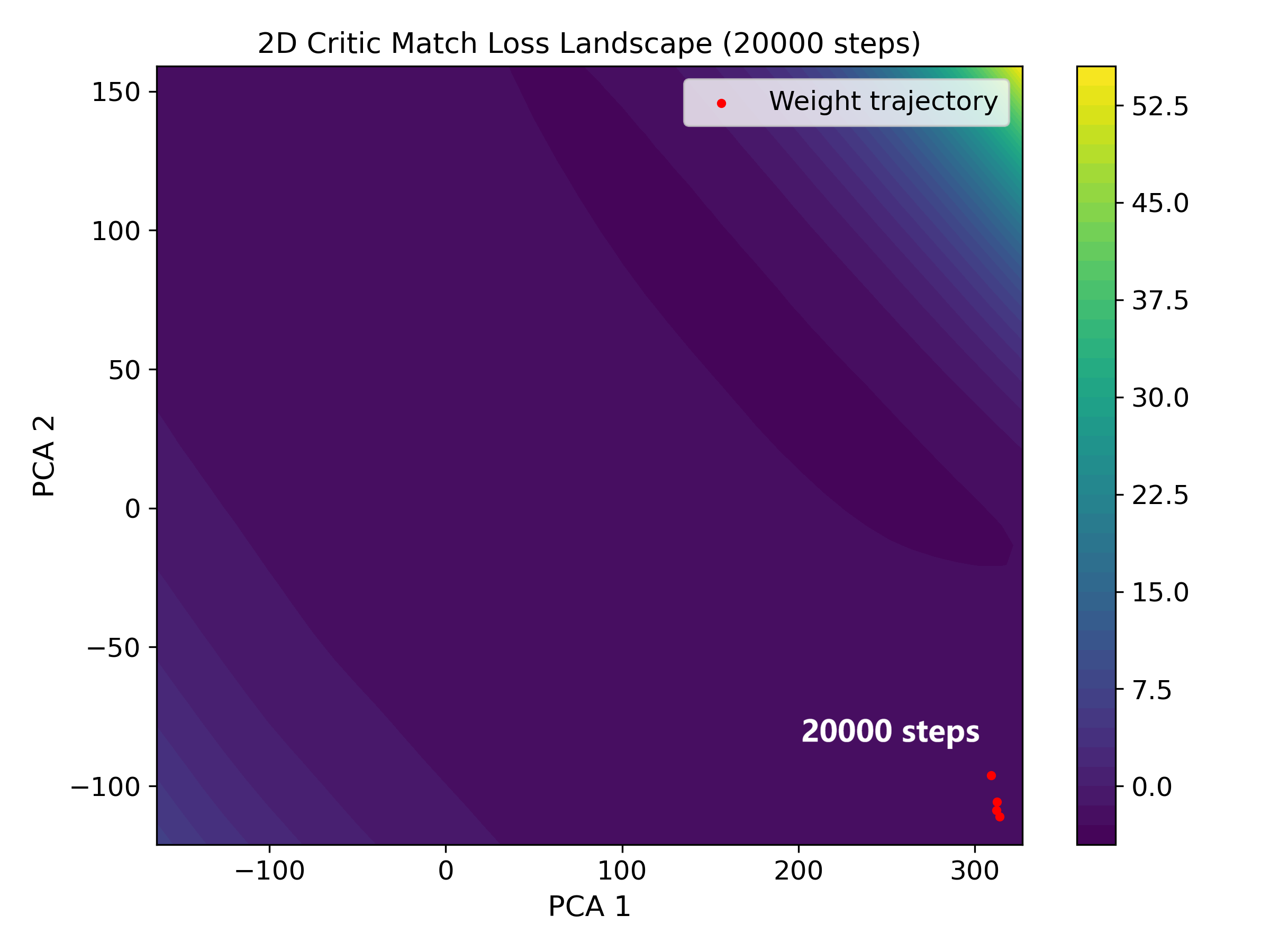}
    \label{fig:SAC_2d_step20000}%
}

\subfigure[3-D landscape at step 50000]{
    \includegraphics[width=0.48\textwidth]{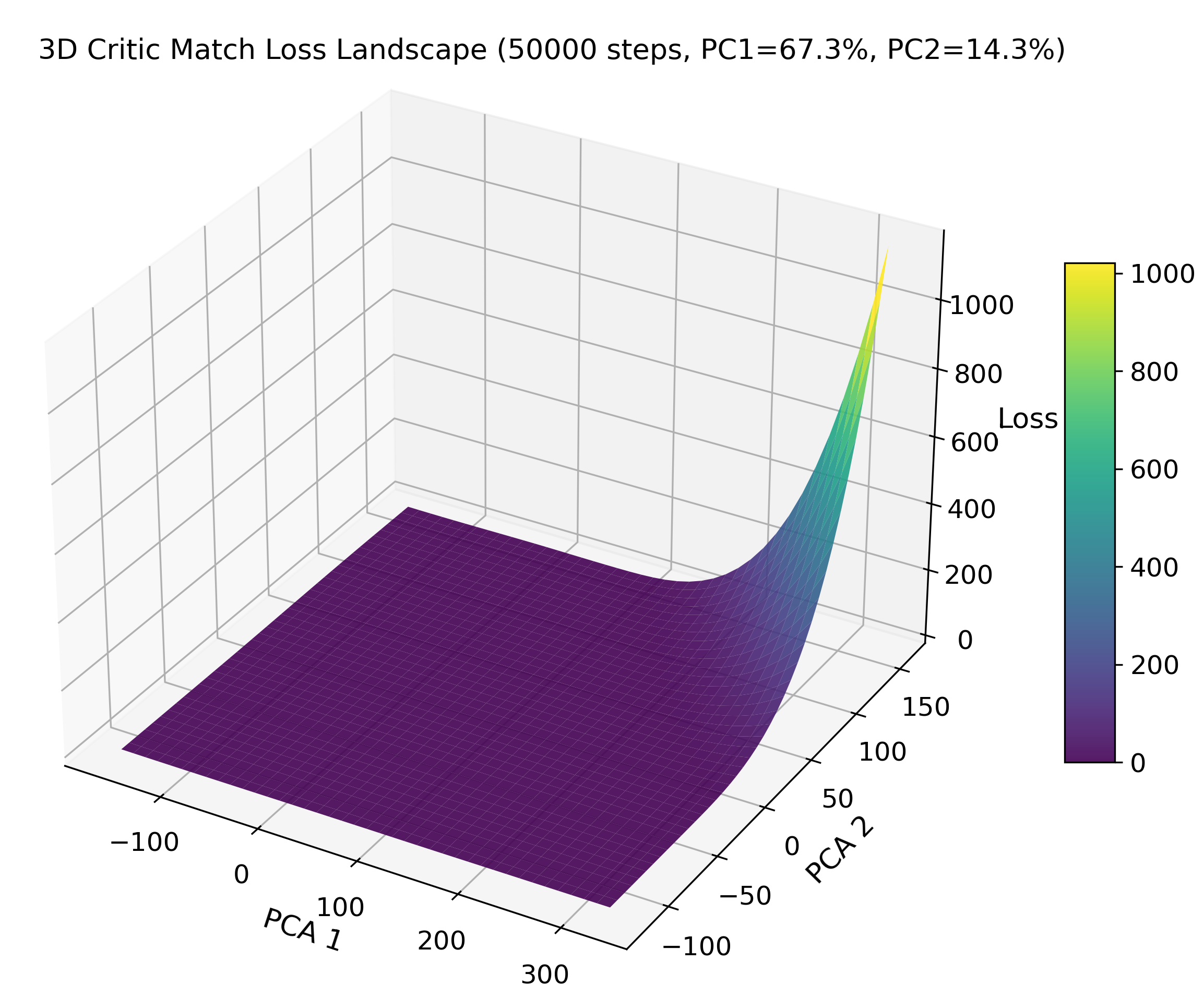}
    \label{fig:SAC_3d_step50000}
}
\hfill
\subfigure[2-D landscape at step 50000]{
    \includegraphics[width=0.48\textwidth]{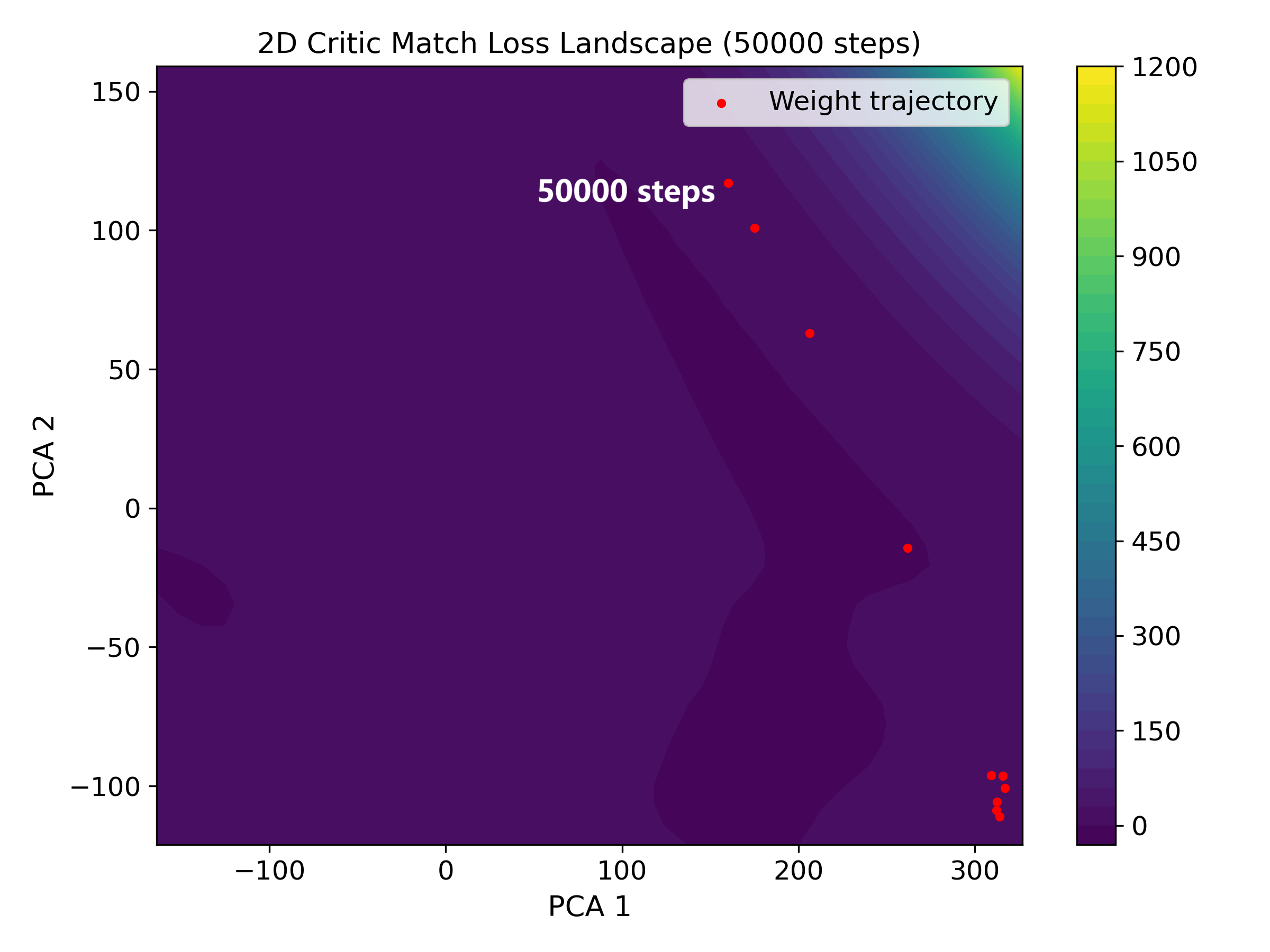}
    \label{fig:SAC_2d_step50000}
}

\subfigure[3-D landscape at step 100000]{
    \includegraphics[width=0.48\textwidth]{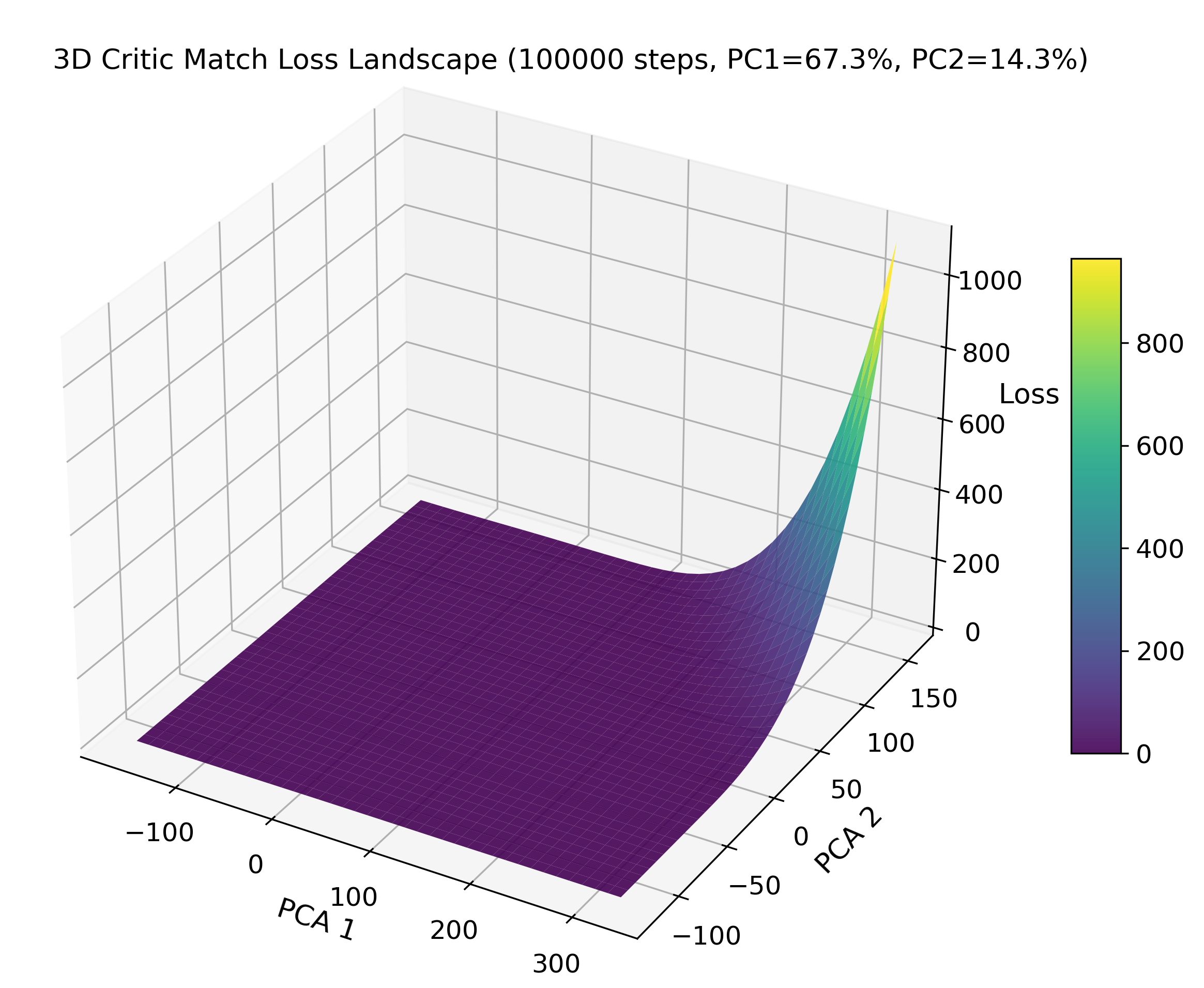}
    \label{fig:SAC_3d_step100000}
}
\hfill
\subfigure[2-D landscape at step 100000]{
    \includegraphics[width=0.48\textwidth]{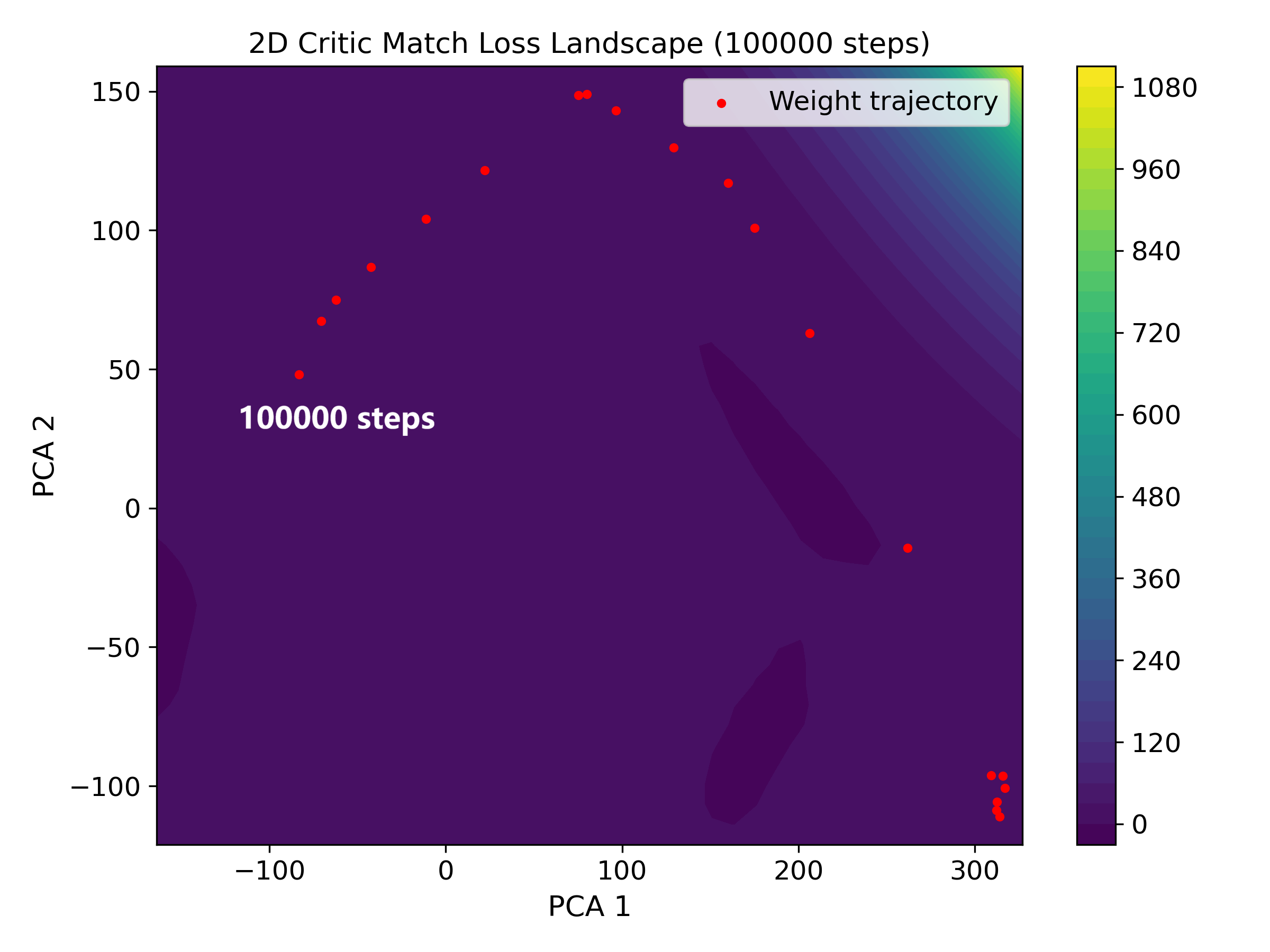}
    \label{fig:SAC_2d_step100000}
}

\caption{Snapshots during training of 3-D and 2-D loss landscape of divergent spacecraft attitude control under SAC with convergent critic}
\label{fig:SAC_spc_loss landscap_div_snapshots}

\end{figure}

To further examine the optimization dynamics, \autoref{fig:SAC_spc_loss landscap_div_snapshots} demonstrates temporal snapshots of the critic match loss landscape which are generated at 20k, 50k, 100k, and 200k steps. The temporal sequence of critic match loss landscapes provides insight into the evolution of critic optimization geometry under the frozen-target approximation.

\autoref{fig:SAC_spc_loss landscap_div_snapshots} further clarify this mechanism. At 20k steps, the trajectory remains concentrated within a compact region of the PCA plane, indicating coherent parameter refinement. Around 50k steps, following the first major critic loss spike in the scalar critic loss curve, the projected trajectory expands and separates into distant regions, suggesting a substantial parameter relocation. At 100k and 200k steps, although the overall 3D surface topology remains qualitatively similar, the trajectory remains fragmented across distinct clusters. The geometry is largely preserved, but the optimization path becomes discontinuous.By 200k steps, the clustering pattern persists. The final critic parameters locate within one geometrically structured region that appears locally stable under the frozen objective, yet remains disconnected from earlier regions explored during training. The trajectory history therefore does not represent a continuous refinement toward a single low-loss basin, but rather a sequence of regime shifts across the frozen-target geometry.

An important observation is that the global 3D surface topology does not change dramatically across temporal snapshots, Under the frozen-target approximation, even when the closed-loop behavior ultimately fails, the critic often evolves within geometrically structured and locally stable regions. This suggests that the divergent SAC case may correspond to a geometrically stable yet ineffective control solution at multiple stages of training.

These observations indicate that, in the case of divergent SAC control with convergent critic parameters, closed-loop failure does not stem from an ill loss surface, but from unstable critic optimization dynamics developed as trajectory discontinuities. The critic match loss landscape therefore provides geometric context for understanding optimization stability, rather than serving as a direct predictor of control convergence.

\subsection{Comparison between convergent and divergent SAC control}
A direct comparison among the convergent SAC case and the two divergent SAC cases reveals several important distinctions in critic optimization geometry.

First, the static 3D landscape alone cannot serve as a definitive indicator of closed-loop success. In both the convergent SAC case and the divergent SAC case with a smooth critic geometry, the final frozen-target landscape exhibits broad smooth regions and coherent basin-like structures. The global topology does not appear pathological in either case. This confirms that a geometrically structured critic surface does not guarantee successful control.

In the divergent SAC case with divergent critic parameters, the key geometric feature is strong directional domination. The PCA analysis shows that nearly all variance is captured by the first principal component, indicating that critic updates are effectively confined to a single dominant direction. The 3-D surface resembles an elongated ridge rather than a balanced basin, and the projected trajectory extends predominantly along this direction. This directional phenomenon is not evident from the scalar weight-norm curve alone. Only through PCA variance ratios and 2-D projected trajectories does the underlying instability mechanism become visible.

In contrast, the divergent SAC case with convergent critic geometry exhibits a smooth final surface without pronounced anisotropy. However, the projected trajectory reveals discontinuous parameter relocations across separated regions of the PCA plane. Here, instability is not caused by geometric degeneration of the surface itself, but by abrupt transitions in optimization dynamics. These transitions may be associated with interactions between critic updates and other components of the actor–critic architecture, including policy saturation effects, exploration dynamics, reward scaling, or replay-induced distribution shifts, which require further investigation.

This distinction becomes clearer when examining weight dispersion in the PCA plane. In the convergent SAC case, dispersion is moderate and progressively concentrates within a coherent low-loss region, reflecting stable refinement. In the divergent SAC case with convergent critic geometry, dispersion appears as fragmented clusters, consistent with discontinuous parameter relocations across separated regions. In the divergent SAC case with divergent critic parameters, dispersion is not clustered but directionally amplified along a dominant principal component, revealing anisotropic instability. In contrast, the divergent ADHDP case exhibits narrow dispersion, suggesting limited weight exploration rather than convergence. These comparisons indicate that dispersion magnitude alone is insufficient, while its geometric structure and relation to trajectory continuity determine its interpretive value.

These observations clarify the interpretation of critic match loss visualization. The method should not be used as a binary indicator of control convergence based solely on surface shape. Instead, it provides a geometric diagnostic framework.By integrating the 3-D loss geometry and the projected 2-D optimization trajectory, the method characterizes the evolution of critic parameters within the loss landscape. It enables the identification of distinct optimization patterns, including dominance by a single unstable direction or fragmentation across disconnected regions. This combined geometric and trajectory analysis clarifies how the critic evolves during training and explains the practical value of the critic match loss landscape.

\subsection {Interpretation scope and methodological boundaries of the frozen-target landscape}
The critic match loss landscape presented in this work is constructed under a frozen-target approximation. By fixing the replay batch and precomputing target values, the dynamic Bellman update process is locally transformed into a stationary surrogate objective. While this simplification is necessary to obtain a consistent and plottable surface, it introduces an inherent gap between the static landscape and the fully non-stationary optimization process encountered during reinforcement learning.

First, the critic match loss evaluated on the frozen-target surface does not coincide exactly with the instantaneous critic loss recorded during training. The true critic loss evolves with changing targets, replay distributions, and policy updates, whereas the match loss is computed under fixed targets associated with a specific policy snapshot. As a result, the weight trajectory projected onto the PCA plane does not correspond one-to-one with the scalar critic loss curve. Nevertheless, the frozen-target landscape remains informative in revealing structural properties of parameter evolution, such as the presence of dominant weight update, clustering behavior, and transitions across regions of varying geometric sensitivity. Even though the absolute loss values differ, these geometric events often align temporally with significant fluctuations in the scalar training curves,

Second, the static landscape does not represent the global loss surface over all possible policies or target configurations. It reflects only the geometry under a specific frozen policy and replay batch. Consequently, it cannot evaluate the critic performance across the full dynamic distribution shift induced by actor–critic coupling. To partially address this limitation, temporal snapshots are introduced. By constructing frozen-target landscapes at multiple training stages, the method captures how the effective local geometry evolves over time. Although each snapshot remains static, the sequence of landscapes provides insight into regime transitions and structural stability during training.

The interpretability strength of this frozen-target geometry depends on the structural complexity of the underlying algorithm. In online methods such as ADHDP, critic updates are tightly coupled with policy behavior and target computation, and relatively few auxiliary mechanisms need to be frozen. Consequently, the frozen-target objective more closely resembles the effective training objective, and geometric instability of the critic is directly reflected in control performance. In contrast, SAC incorporates replay buffering, target networks, twin critics, and entropy regularization, all of which partially decouple critic optimization from immediate closed-loop behavior. As a result, a larger set of components must be frozen to define a stationary surrogate objective, and the geometric stability of the critic under this surrogate does not necessarily imply control convergence. This difference arises primarily from architectural complexity rather than from the freezing procedure itself.

In conclusion, these considerations clarify the interpretation scope of the critic match loss visualization. The static landscape should not be viewed as a direct reconstruction of the full dynamic reinforcement learning objective. Although closed-loop performance is governed by broader interactions within the actor–critic system, the critic match loss landscape serves as a geometric diagnostic tool that characterizes critic optimization behavior at selected training stages. Its value lies in revealing structural properties of the loss geometry, directional dominance in parameter updates, and discontinuous transitions in weight.

\section{Validation on Benchmark System}
To validate the proposed critic match loss landscape visualization method for off-policy reinforcement learning algorithms, the continuous cart-pole system is selected as a benchmark to further demonstrate the interpretation procedure of the method.

The dynamics of the continuous cart-pole system is given in detail in \ref{app:Appendix continuous cart-pole dynamics}. The system was trained using the Soft Actor--Critic (SAC) algorithm from the Stable-Baselines3 library \citep{stable-baselines3}. The actor and critic networks adopt the default multi-layer perceptron (MLP) architecture with twin Q-functions and a stochastic Gaussian policy. Automatic entropy tuning is enabled to adaptively adjust the temperature parameter. The learning rate is set to $3\times10^{-4}$ for all networks. 
Training is performed for $5\times10^{4}$ interaction steps with simulation step size $\ 0.02\,\mathrm{s}$. A replay buffer of size $10^{5}$ stores past transitions, from which mini-batches of $256$ samples are drawn. Learning begins after the first $10^{3}$ collected samples. The discount factor and soft target update coefficient are chosen as $\ 0.99$ and $\ 0.005$, respectively, and one gradient update is executed per environment step.

To encourage stable upright balancing and smooth control, a quadratic cost structure is adopted. 
Let the system state be $s = [\theta, \dot{\theta}, x, \dot{x}]^{\top}$ and the normalized control action be $a$. The instantaneous cost is defined as
\begin{equation}
\begin{aligned}
c(s,a) =\;&
w_{\theta}\theta^{2}
+ w_{\dot{\theta}}\dot{\theta}^{2}
+ w_{x}\left(\frac{x}{x_{\max}}\right)^{2}
+ w_{\dot{x}}\dot{x}^{2}
+ w_{u}a^{2},
\end{aligned}
\end{equation}
where $\theta$ and $\dot{\theta}$ are converted to radians, $x_{\max}$ denotes the track limit, and $w_{\theta}, w_{\dot{\theta}}, w_{x}, w_{\dot{x}}, w_{u}$ are positive weighting coefficients.

In this work, the weights are empirically selected as
\begin{equation}
w_{\theta}=5.0, \quad
w_{\dot{\theta}}=0.1, \quad
w_{x}=1.0, \quad
w_{\dot{x}}=0.1, \quad
w_{u}=0.01.
\end{equation}

The reward is defined as the negative cost,
\begin{equation}
r(s,a) = -c(s,a),
\end{equation}
which encourages the pole to remain upright, suppresses excessive velocities, keeps the cart near the center of the track, and penalizes aggressive control inputs.

Additionally, a large terminal penalty is imposed when safety limits are violated,
\begin{equation}
r_{\text{fail}} = -50,
\end{equation}
if $|x| > x_{\max}$ or $|\theta| > \theta_{\max}$. 
This extra penalty provides a strong learning signal to discourage failure states.

\begin{figure*}[htbp]
    \centering
    \subfigure[SAC control cart-pole result]{
        \centering
        \includegraphics[width=0.6\textwidth]{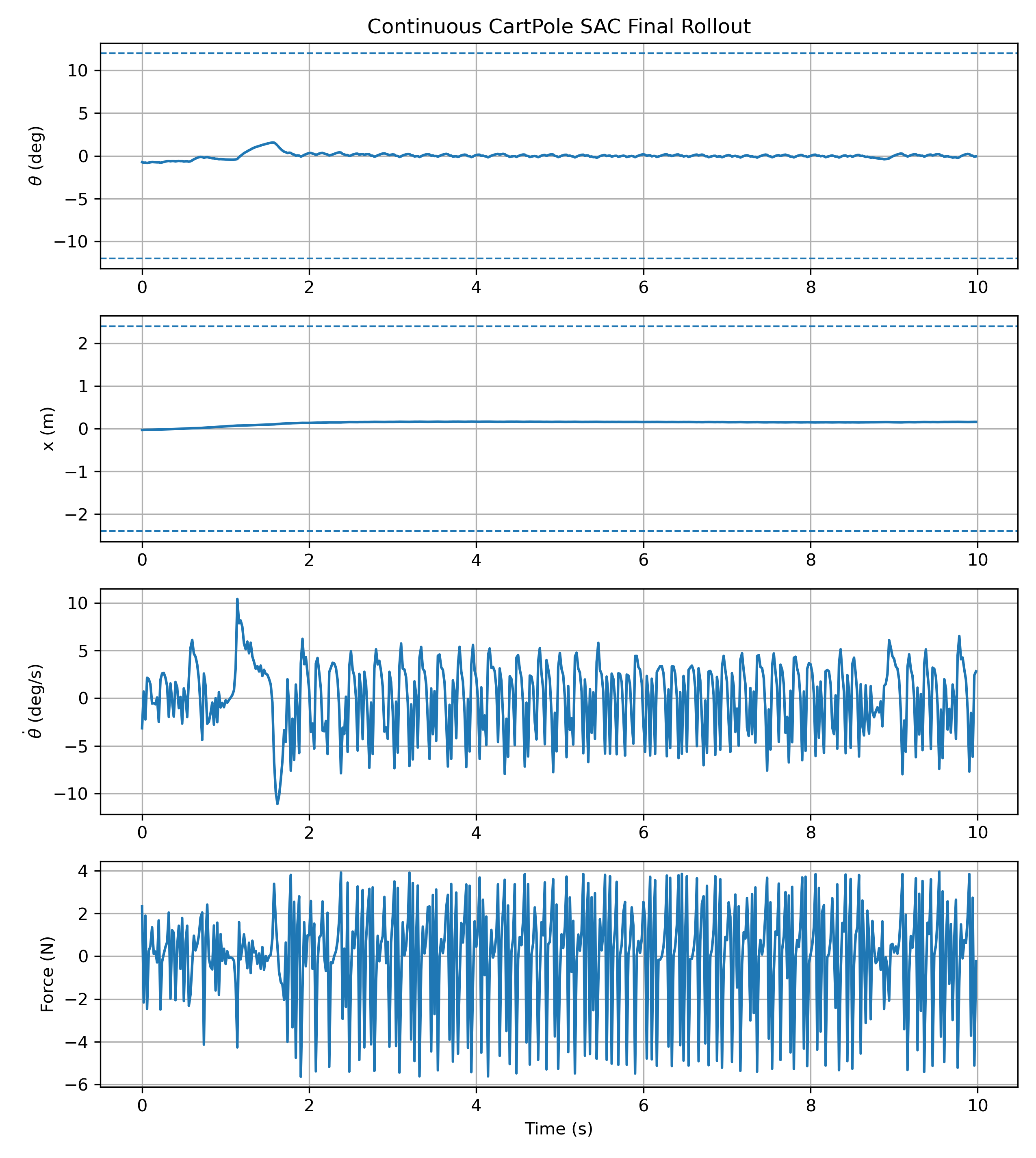}
        \label{fig: SAC cart-pole system performance}
}
    \hspace{0.02\columnwidth}
    \subfigure[SAC cart-pole training curve]{
        \centering
        \includegraphics[width=0.6\textwidth]{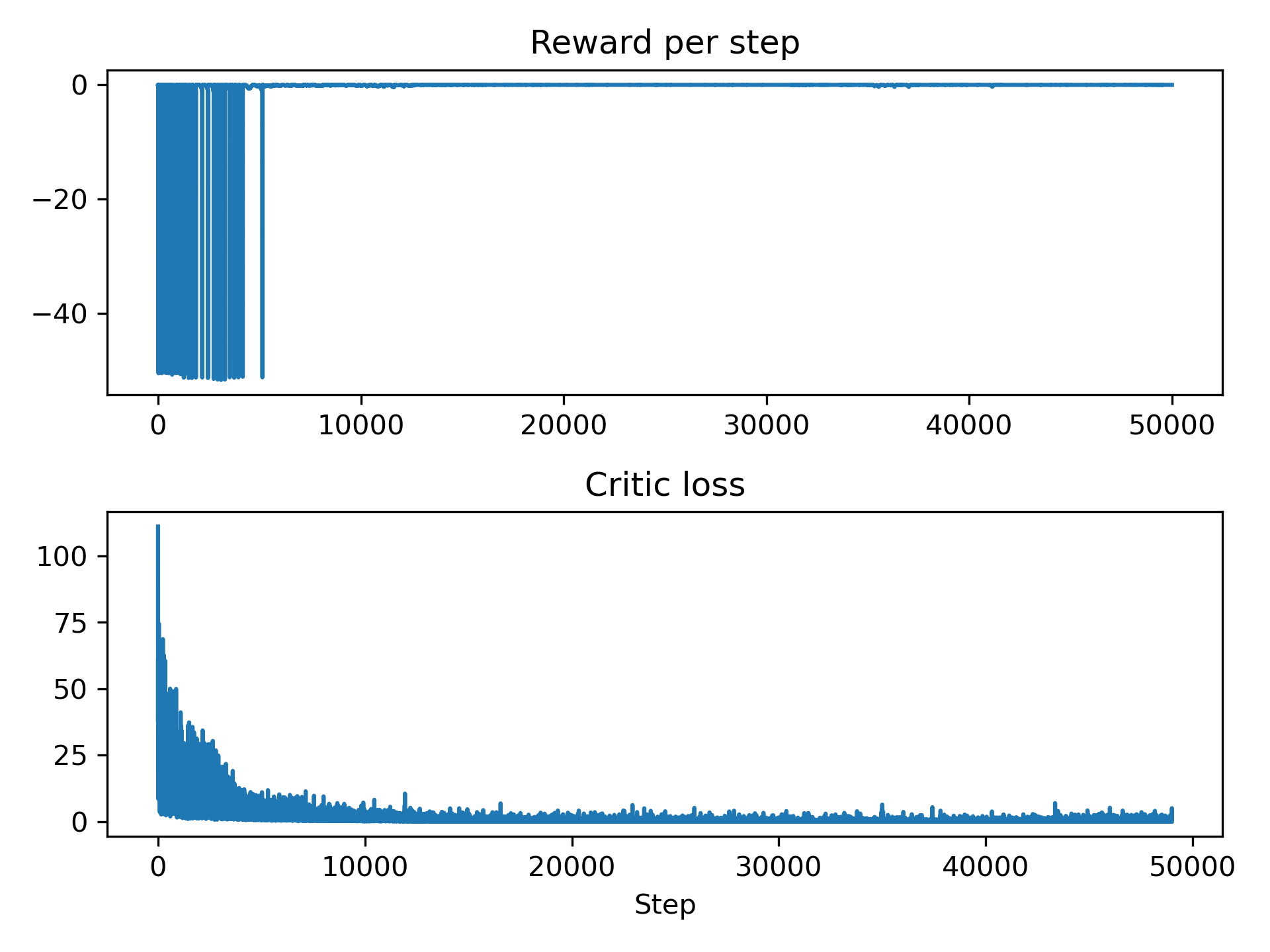}
        \label{fig: SAC cart-pole training curve}
}

    \caption{SAC control results for cart-pole system}
    \label{fig:SAC cart-pole system and algorithm performance}
\end{figure*}

After training, the learned policy is evaluated using a deterministic rollout of $500$ simulation steps. 
The resulting pole angle, angular velocity, cart displacement, applied control forces and the training curves are shown in \autoref{fig:SAC cart-pole system and algorithm performance}.

The training curves of the SAC algorithm on the continuous cart-pole system are shown in \autoref{fig: SAC cart-pole training curve}. The reward rapidly increases from large negative values and stabilizes near zero, indicating successful pole stabilization and cart position regulation. Meanwhile, the critic loss decreases monotonically during the early training phase and gradually converges to a small and stable range without significant oscillations. This behavior suggests stable critic approximation and consistent Bellman error minimization.

The final rollout of the cart-pole system further confirms successful control performance. The pole angle remains close to the upright equilibrium, the cart position stays within bounded limits, and the control force remains finite and well-regulated. These results demonstrate that the SAC algorithm achieves stable convergence in this benchmark environment

\begin{figure}[htbp]
    \centering
    \subfigure[SAC cp 3-D loss landscape]{
        \centering
        \includegraphics[width=0.5\textwidth]{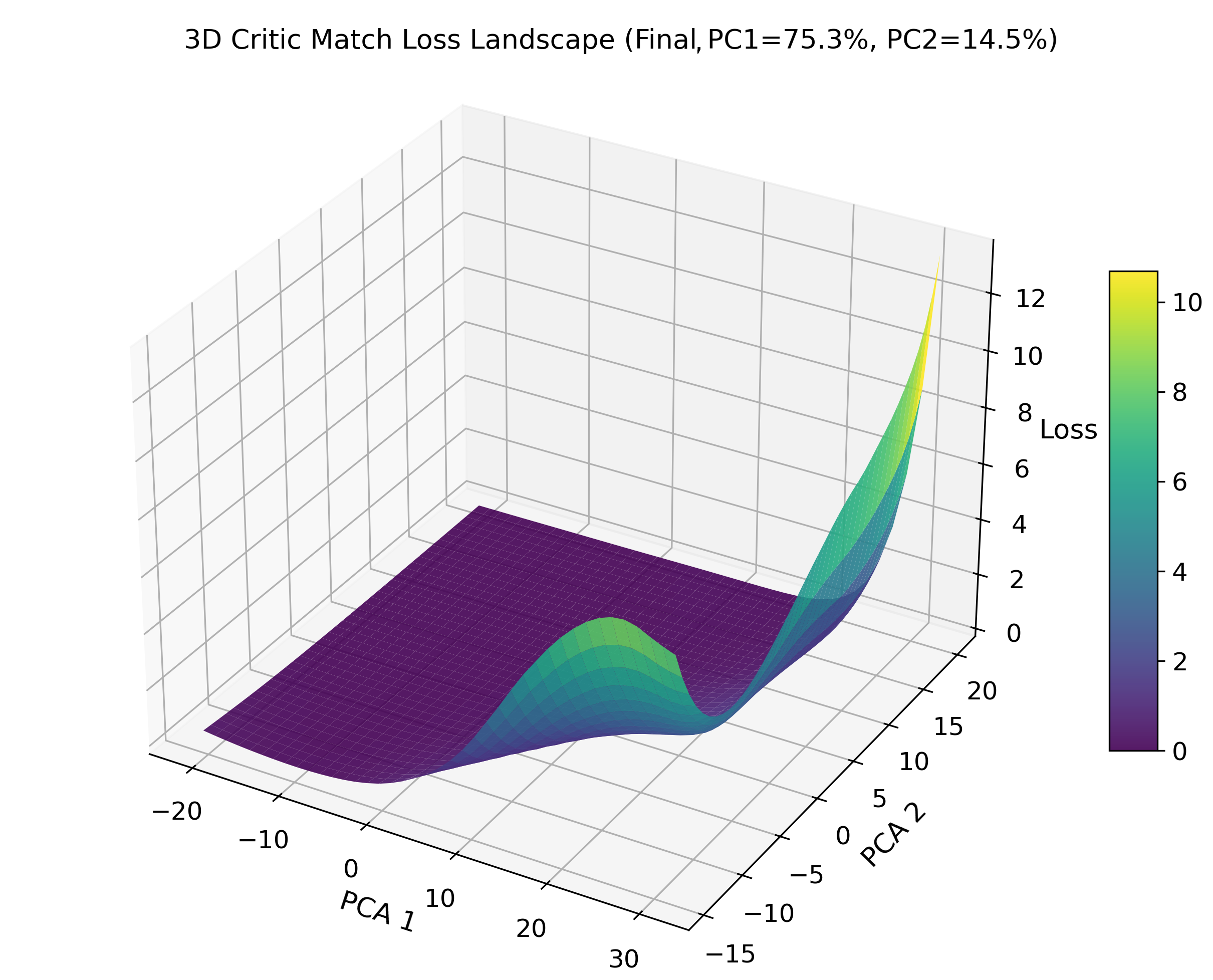}
        \label{fig:SAC_3D loss cp}
    }
    \hspace{0.02\columnwidth}
    \subfigure[SAC cp 2-D loss curve]{
        \centering
        \includegraphics[width=0.5\textwidth]{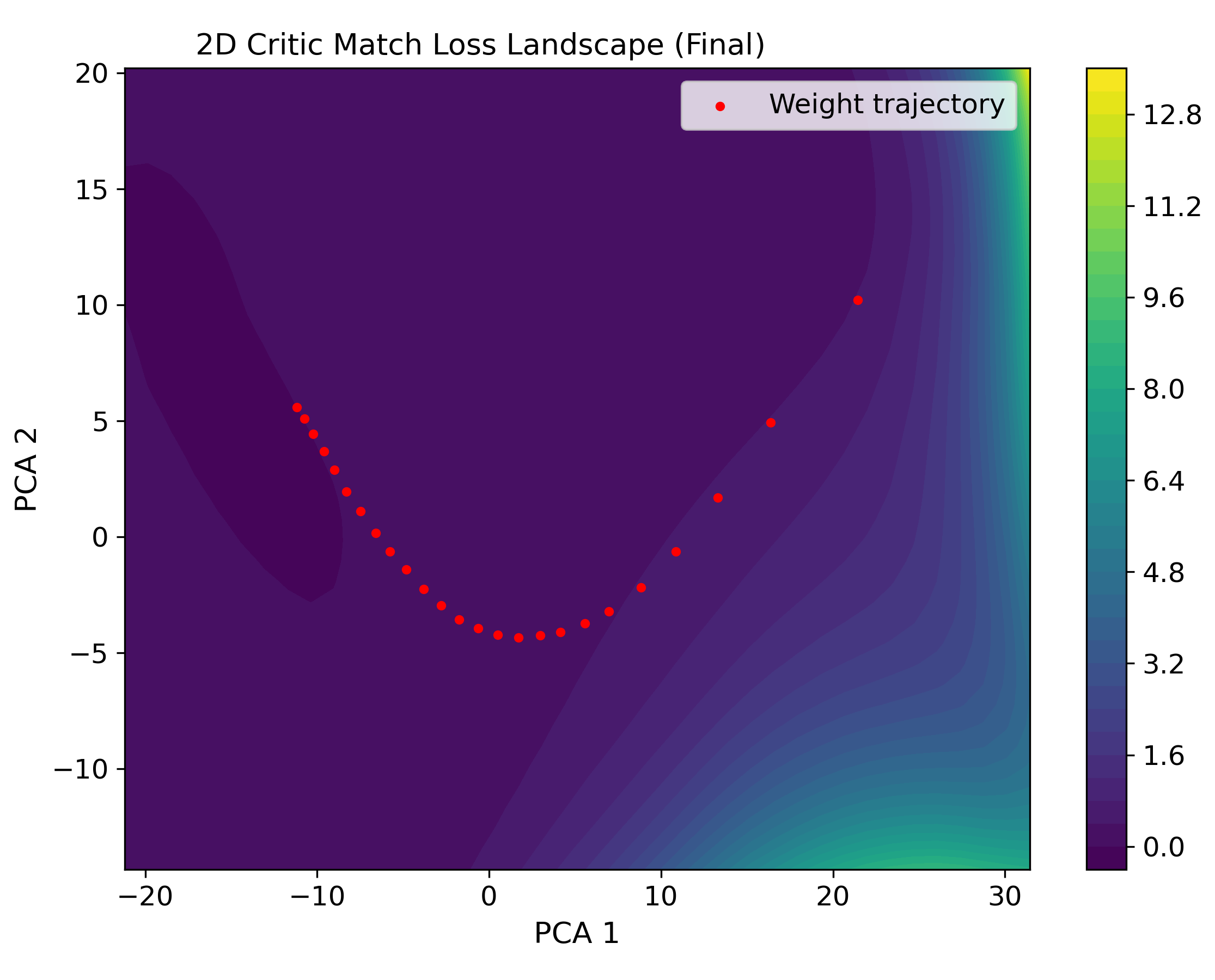}
        \label{fig:SAC_2D loss cp}
}
    \caption{3-D and 2-D loss landscape of spacecraft attitude SAC control}
    \label{fig:SAC_cp_loss landscape}
\end{figure}

The corresponding critic match loss landscape is presented in \autoref{fig:SAC_cp_loss landscape}. Similar to the convergent spacecraft case, the loss surface exhibits a clear concave valley structure with a well-defined low-loss basin. When projected onto the PCA plane, the recorded critic weight trajectory exhibits a gradual progression toward the low-loss region defined by the frozen-target objective, with the final parameters situated inside the basin. The surface curvature is smooth and does not display irregular spikes or fragmented contour patterns.

Although the cart-pole system differs significantly from spacecraft attitude dynamics in state dimension and physical structure, the geometric characteristics of the critic match loss landscape remain consistent with those observed in the convergent spacecraft case. This consistency suggests that the relationship between landscape geometry and critic convergence is not specific to a particular dynamical system, but reflects a more general property of stable off-policy actor–critic training.

\section{Conclusion}
Off-policy reinforcement learning has a different structure from online reinforcement learning in data flow and target calculation in TD error. This work adapts the critic match loss landscape visualization method originally developed for online reinforcement learning to the off-policy reinforcement learning framework by explicitly addressing differences in data flow and target construction. The Soft Actor–Critic (SAC) algorithm is adopted as the demonstration case of the adapted critic match loss landscape method for off-policy RL. By applying SAC to the spacecraft attitude control problem, the critic match loss landscape is reconstructed and compared with that of the ADHDP algorithm using sharpness, basin area, and local anisotropy metrics, together with temporal landscape snapshots. The results show that the convergent SAC case exhibits a coherent basin structure under the frozen-target objective, while the ADHDP case presents sharper and more irregular geometry. The divergent SAC case further demonstrates that geometric stability of the critic under the surrogate objective does not necessarily guarantee closed-loop convergence, reflecting the influence of broader algorithmic interactions. Overall, the proposed visualization framework serves as a geometric diagnostic tool for interpreting critic optimization behavior. Although it does not reconstruct the full non-stationary training objective, it provides a consistent approach for analyzing structural properties of critic learning across both online and off-policy actor–critic algorithms.

\appendix
\section{Attitude dynamics of spacecraft with unknown inertia}
\label{app:Appendix attitude dynamics}

In this paper, a body-fixed reference frame $\mathcal{B}$ is defined at the center of mass of the combined spacecraft, with its axes aligned to the principal directions of inertia. An Earth-Centered Inertial (ECI) frame $\mathcal{N}$ is also defined, with unit vectors $\{\bm{\overline{n_1}},\bm{\overline{n_2}},\bm{\overline{n_3}}\}$. The $\bm{\overline{n_1}}$ axis points toward the mean equinox, the $\bm{\overline{n_3}}$ axis is parallel to the Earth's rotation axis (celestial north), and $\bm{\overline{n_2}}$ completes the right-handed system.

The attitude of the combined spacecraft is represented by a unit quaternion 

\begin{equation}
\bm{q} = 
\begin{bmatrix}
q_0 \\
q_1 \\
q_2 \\
q_3
\end{bmatrix}
\end{equation}

which defines the rotation from $\mathcal{N}$ to $\mathcal{B}$. This formulation avoids singularities and is suitable for cases where the attitude may vary over a wide range. The angular velocity of the spacecraft expressed in frame $\mathcal{B}$ is denoted as 

\begin{equation}
\bm{\omega} =
\begin{bmatrix}
\omega_1 \\
\omega_2 \\
\omega_3
\end{bmatrix}.
\end{equation}

The kinematic equation is 
\begin{equation}
\dot{\bm{q}} = \frac{1}{2} \, \bm{q} \otimes 
\begin{bmatrix}
0 \\
\bm{\omega}
\end{bmatrix}
\label{eq:sc_kinematics}
\end{equation}
where $\otimes$ denotes the quaternion multiplication.

The attitude dynamics are expressed as
\begin{equation}
\bm{M} = \hat{J}_{\mathrm{sc}} \cdot \dot{\bm{\omega}} + \bm{\omega} \times \left( \hat{J}_{\mathrm{sc}} \cdot \bm{\omega} \right)
\label{eq:sc_dynamics}
\end{equation}
Here, $\bm{M}$ is the control torque, and $\hat{J}_{\mathrm{sc}}$ is the inertia matrix of the combined spacecraft, which is assumed to be unknown.

\section{Dynamics of the Continuous Cart-Pole System}
\label{app:Appendix continuous cart-pole dynamics}
A cart of mass $m_c$ moves horizontally along a track, and an inverted pendulum of mass $m$ is attached to the cart by a frictional pivot. The pendulum is constrained to rotate only in the plane. An idealized model is assumed in which the pendulum mass is concentrated at a single point located at the end of a rigid massless rod of length $l$.

The upright configuration corresponds to an unstable equilibrium. Without active control, gravity causes the pendulum to fall away from the vertical position. Therefore, an external horizontal force must be continuously applied to the cart in order to stabilize the pendulum.  In contrast to the classical discrete cart-pole benchmark, where the control input takes values from a finite set, the present formulation adopts a \emph{continuous control action}. Specifically, the controller can generate any real-valued force within a bounded interval, leading to smoother and more realistic actuation.

The system state is defined as
\begin{equation}
s = [\theta,\ \dot{\theta},\ x,\ \dot{x}]^{\top},
\end{equation}
where $\theta$ denotes the angular displacement of the pole from the upright vertical direction, $\dot{\theta}$ is the angular velocity, $x$ is the horizontal displacement of the cart, and $\dot{x}$ is the cart velocity.

The control input is a continuous scalar action
\begin{equation}
a \in [-1,1],
\end{equation}
which is linearly mapped to the physical force applied to the cart,
\begin{equation}
f_{\mathrm{cp}} = a F_{\max},
\end{equation}
where $F_{\max}$ is the maximum allowable force magnitude.

Based on Newtonian mechanics, the nonlinear dynamics of the cart--pole system are described by
\begin{equation}
\ddot{\theta} =
\frac{
g \sin\theta
+
\cos\theta
\left(
\frac{-f_{\mathrm{cp}} - m l \dot{\theta}^{2} \sin\theta}{m_c + m}
\right)
}{
l\left(
\frac{4}{3} - \frac{m \cos^{2}\theta}{m_c + m}
\right)
},
\label{eq:cont_theta}
\end{equation}

\begin{equation}
\ddot{x} =
\frac{
f_{\mathrm{cp}} + m l \left(
\dot{\theta}^{2}\sin\theta - \ddot{\theta}\cos\theta
\right)
}{
m_c + m
},
\label{eq:cont_x}
\end{equation}
where $g=-9.8\,\mathrm{m/s^{2}}$ denotes gravitational acceleration. 
Optional viscous or Coulomb friction terms for both the cart and the pendulum can be incorporated to account for mechanical dissipation.

The control objective is to learn a policy that continuously generates forces to maintain $\theta \approx 0$ (upright balance), suppress angular and translational velocities, keep the cart position within track limits and
avoid excessive control effort. This continuous-action formulation enables smoother control signals and provides a more faithful representation of practical control systems compared with discrete force switching.

\newpage
\section{Control Results of ADHDP for the Spacecraft Attitude System}
\label{app:Appendix ADHDP control results}
\begin{figure}[H]
    \centering
    \subfigure[Quaternion]{
        \centering
        \includegraphics[width=0.6\textwidth]{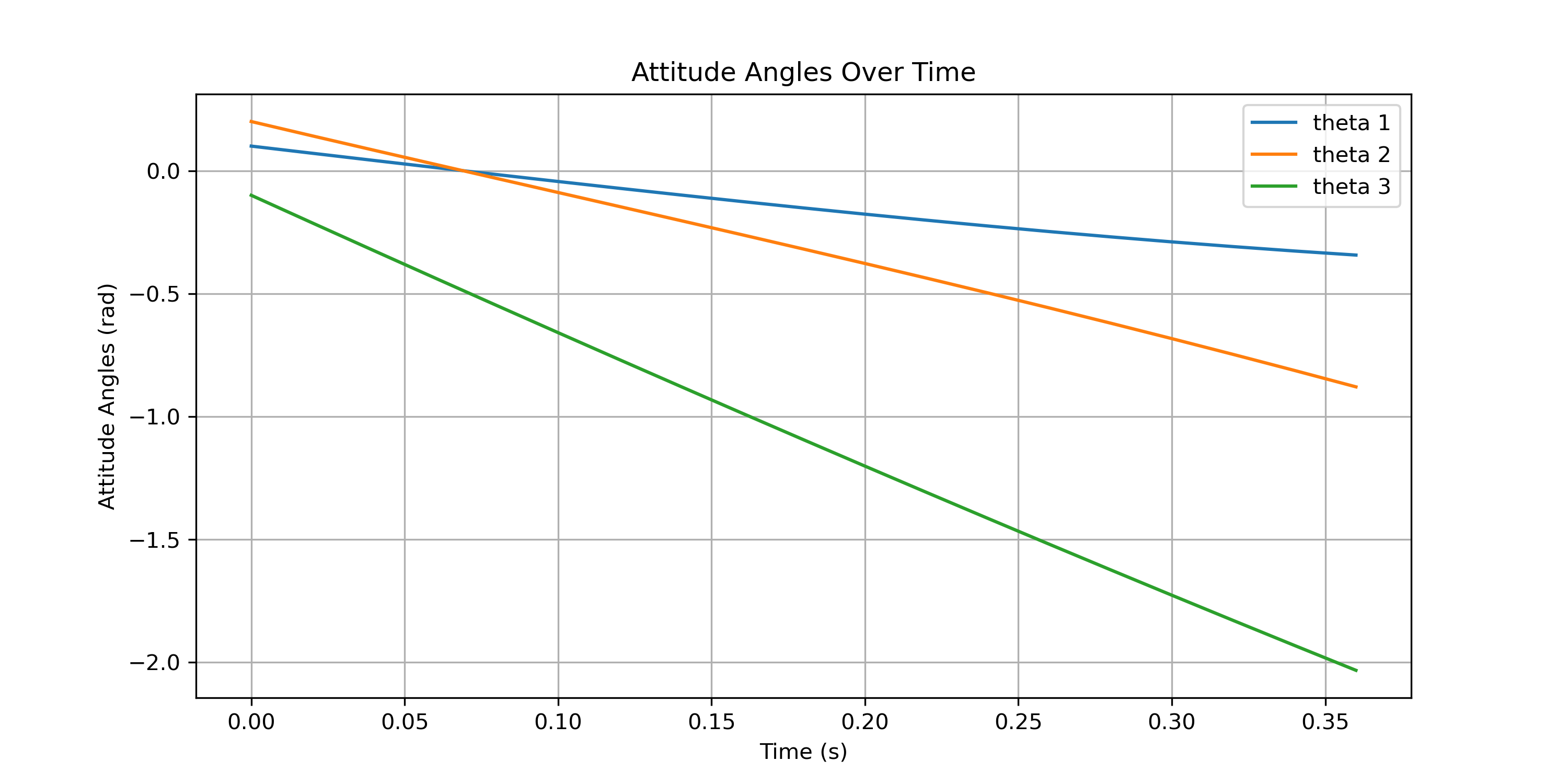}
        \label{fig: ADHDP_Attitude angle}
}
    \hspace{0.01\columnwidth}
        \subfigure[Angular velocity]{
        \centering
        \includegraphics[width=0.6\textwidth]{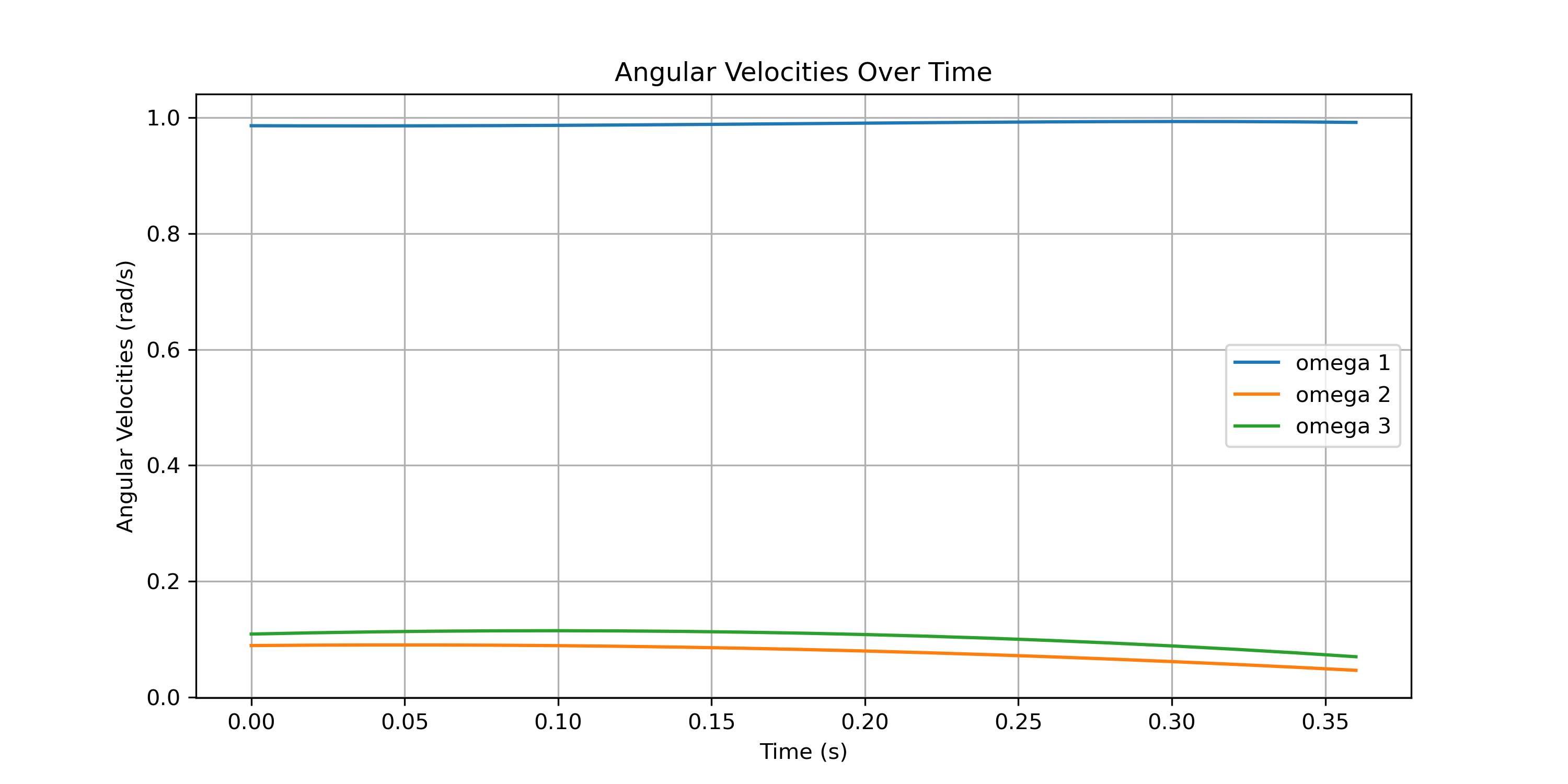}
        \label{fig: ADHDP_Angular velocity}
}

    \hspace{0.01\columnwidth}
        \subfigure[Control torque]{
        \centering
        \includegraphics[width=0.6\textwidth]{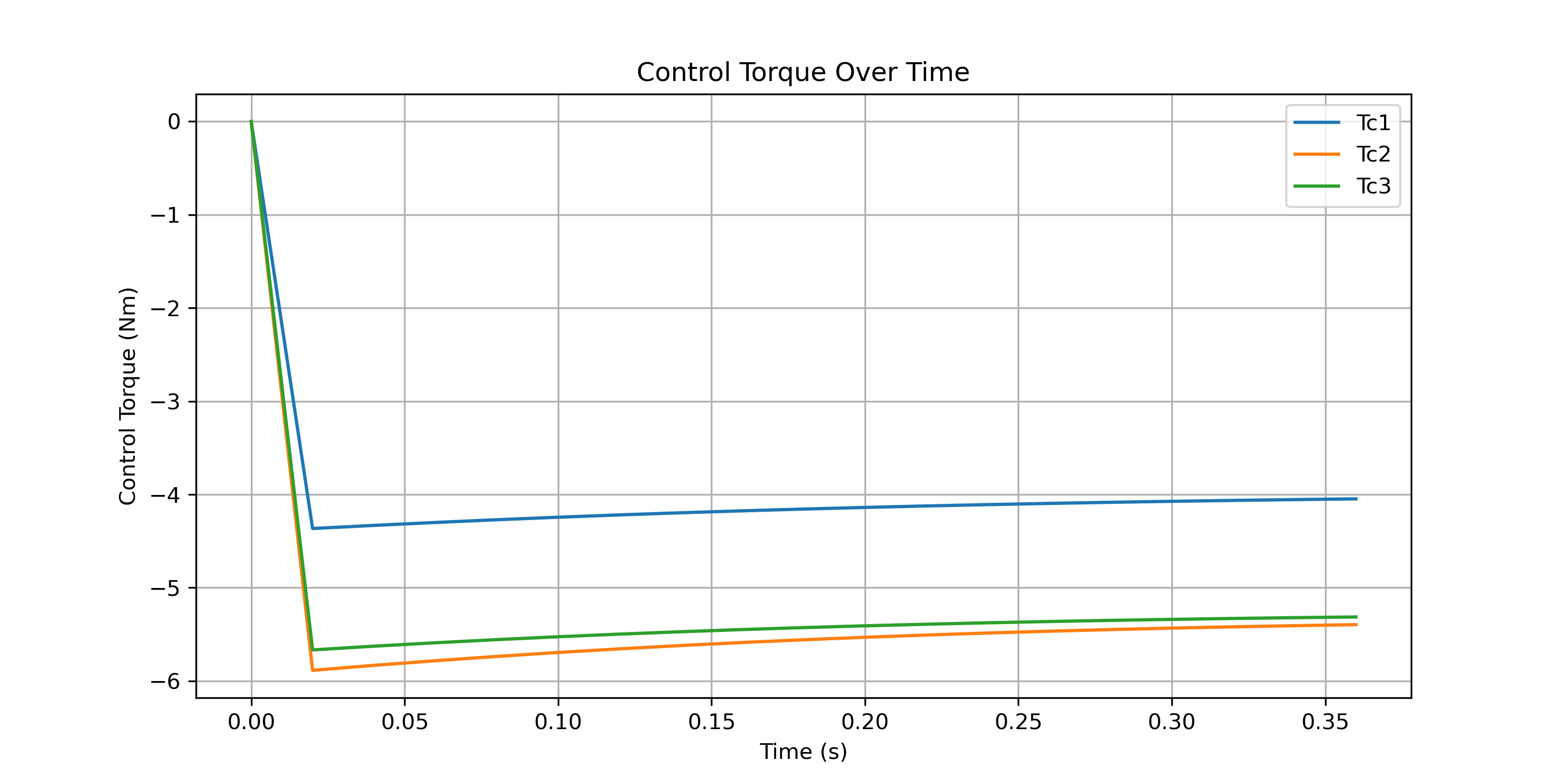}
        \label{fig: ADHDP_Control torque}
}
    \hspace{0.01\columnwidth}
    \subfigure[ADHDP critic training curve]{
        \centering
        \includegraphics[width=0.48\textwidth]{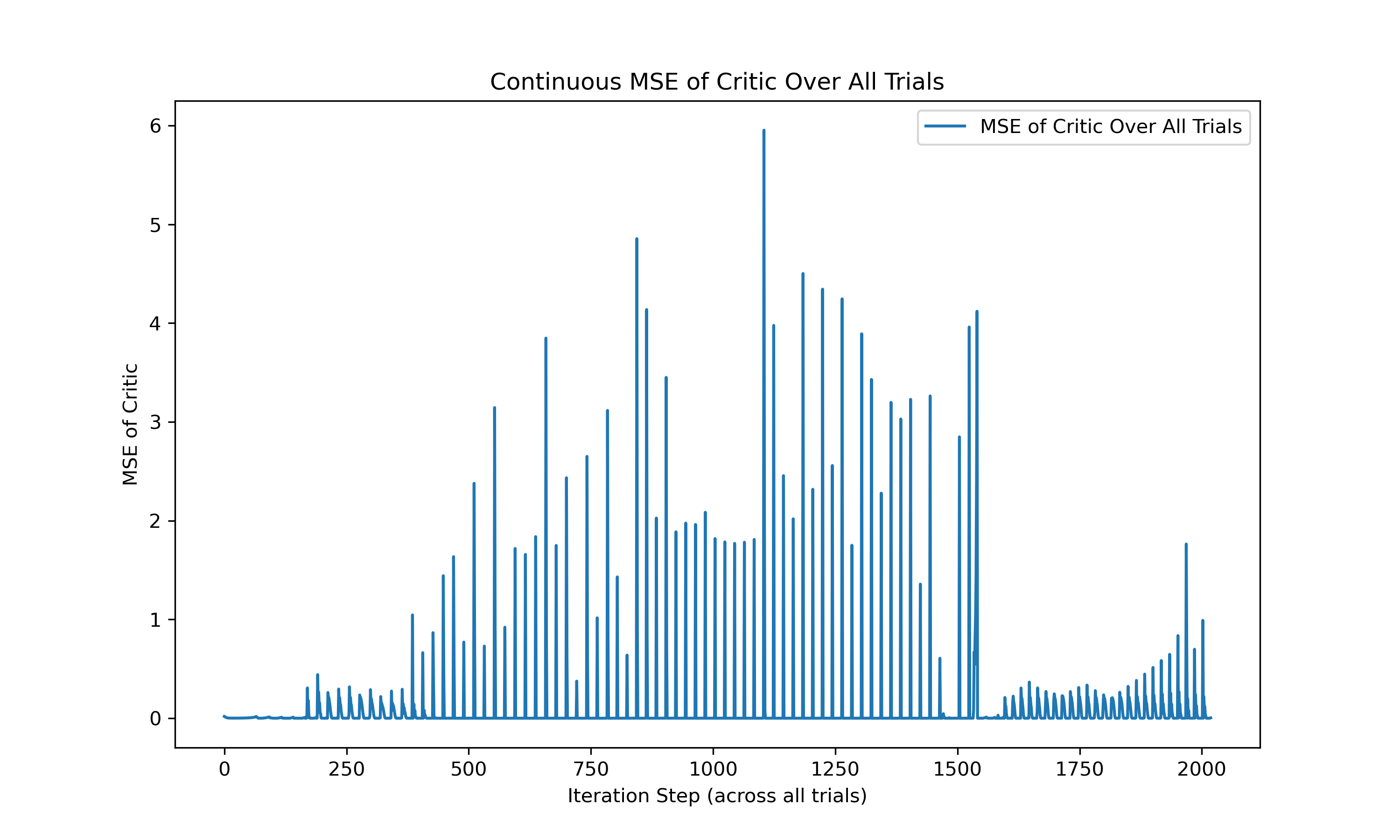}
        \label{fig: ADHDP critic training curve}
}
    \hfill
    \subfigure[ADHDP actor training curve]{
        \centering
        \includegraphics[width=0.48\textwidth]{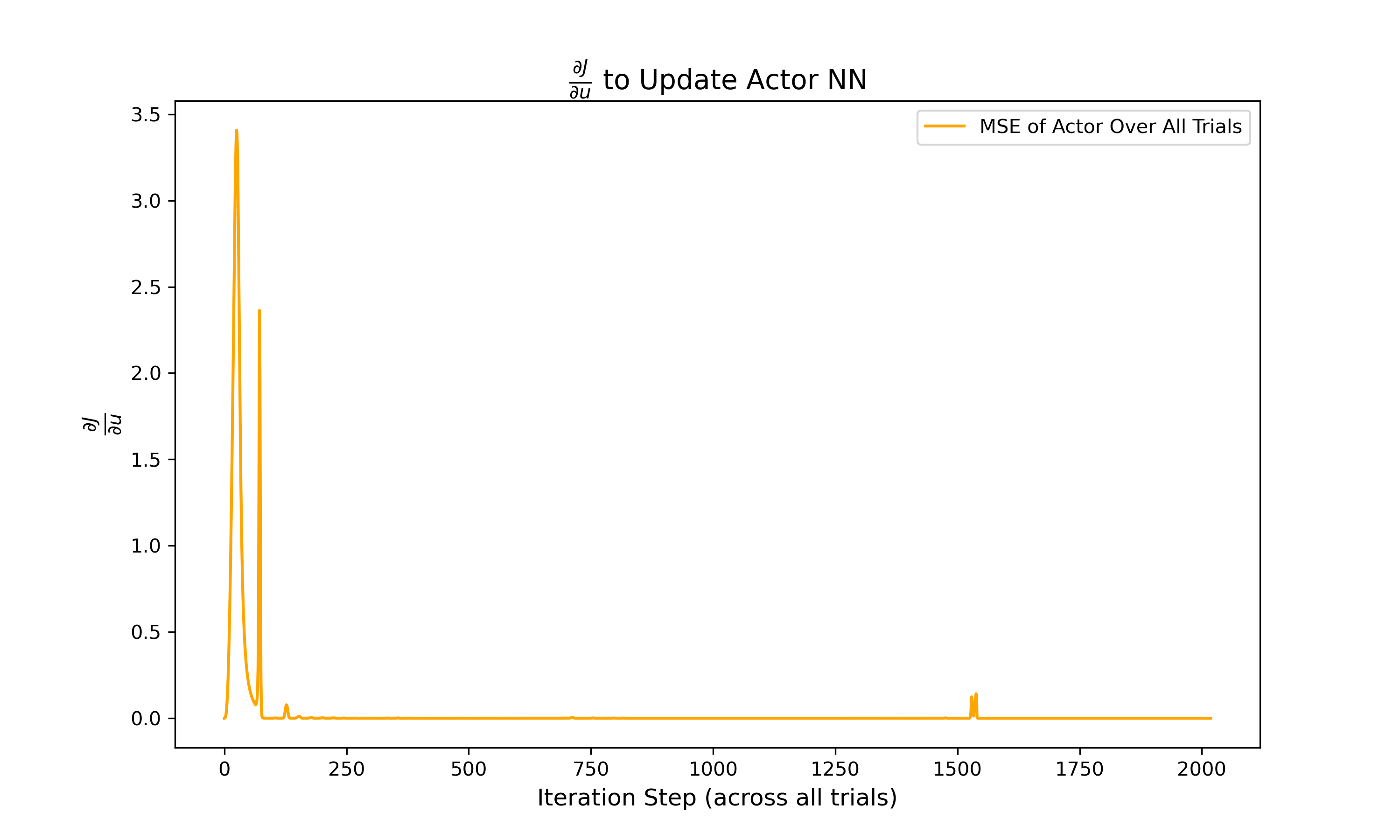}
        \label{fig: ADHDP actor training curve}
}

    \caption{ADHDP control results for cart-pole system}
    \label{fig:ADHDP system and algorithm performance}
\end{figure}



\subsection*{Declaration of competing interest}

The authors have no competing interests to declare that are relevant to
the content of this article.

\section*{References}

The list of references should only include works that are cited in the
text and that have been published or accepted for publication. Personal
communications and unpublished works should only be mentioned in the
text. Papers accepted for publication are cited as ``in press'' and their DOIs. Do not
use footnotes or endnotes as a substitute for a reference list.

\bibliographystyle{astrobib}
\bibliography{maintext}

\end{document}